\documentclass{article}
\usepackage[margin=1.0in]{geometry}
\usepackage{amsmath}
\usepackage{amssymb}
\usepackage[dvipsnames]{xcolor}
\usepackage{natbib}
\usepackage{graphicx} %
\usepackage{caption}
\usepackage{subcaption}
\usepackage{hyperref}
\usepackage{booktabs}
\usepackage{array}
\usepackage{color,soul}

\usepackage{ulem} %

\usepackage{authblk} %

\DeclareMathOperator*{\argmax}{arg\,max}
\DeclareMathOperator*{\argmin}{arg\,min}

\title{Automating the Search for Artificial Life with Foundation Models}

\makeatletter
\renewcommand\AB@affilsepx{, }
\makeatother

\author[1,2]{\mbox{Akarsh Kumar}}
\author[3]{\mbox{Chris Lu}}
\author[4]{\mbox{Louis Kirsch}}
\author[2]{\mbox{Yujin Tang}}
\author[5]{\mbox{Kenneth O. Stanley}}
\author[1]{\mbox{Phillip Isola}}
\author[2]{\mbox{David Ha}}
\affil[1]{MIT}
\affil[2]{Sakana AI}
\affil[3]{OpenAI}
\affil[4]{The Swiss AI Lab IDSIA}
\affil[5]{Independent}

\date{}

\begin{document}

\maketitle

\vspace{-10mm}
\begin{figure}[!h]
    \centering
    \includegraphics[width=0.97\textwidth]{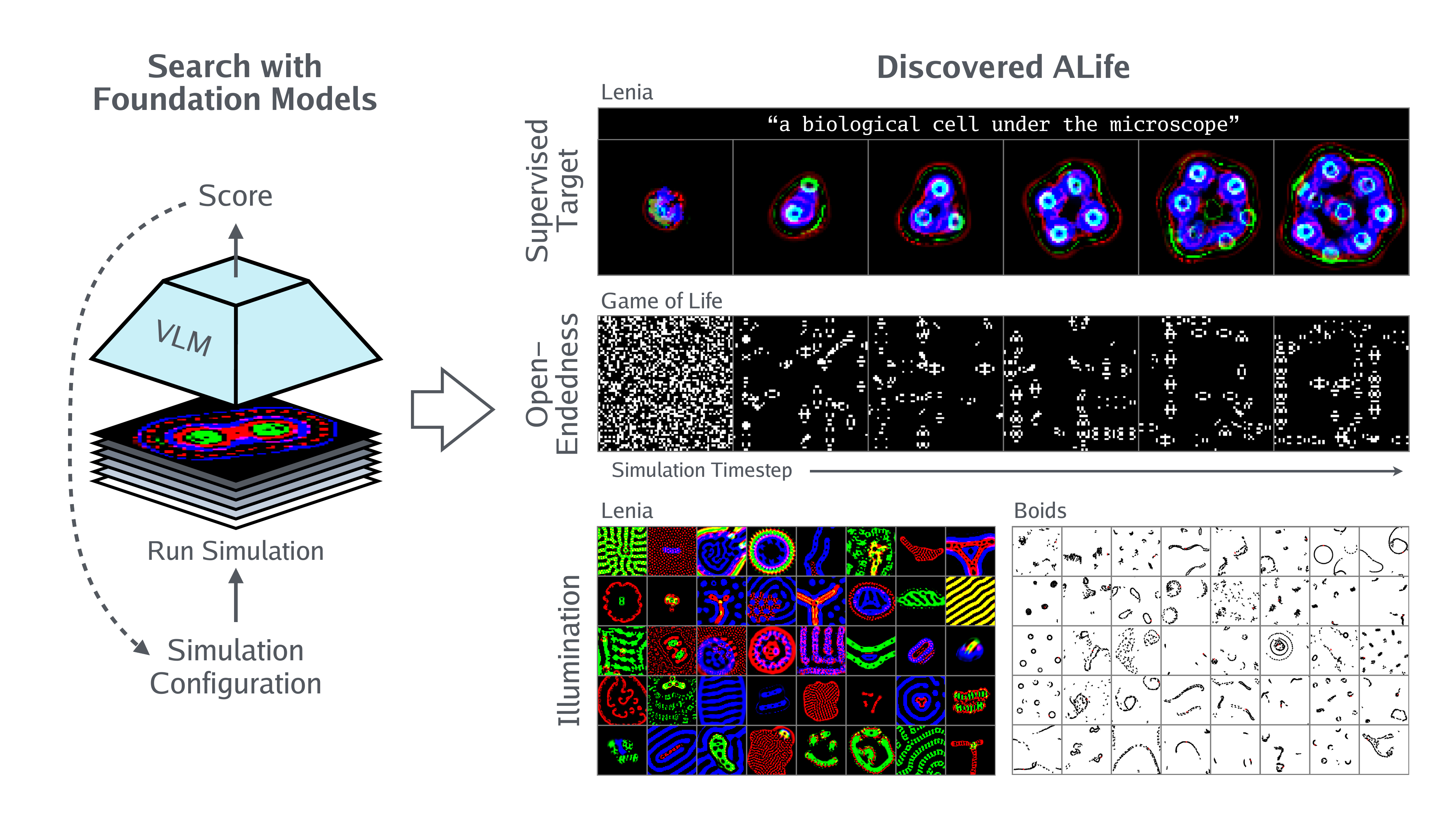} %
    \caption{
    \textbf{Overview:}
    Our method, ASAL, searches for interesting ALife simulations by using a vision-language foundation model to evaluate the simulation's produced videos.
    ALife lifeforms are discovered across different substrates with three different mechanisms: (1) found via a text prompt, (2) found via searching for open-ended simulations, and (3) illuminating a set of diverse simulations.
    }
    \label{fig:teaser}
\end{figure}

\vspace{-3mm}
\begin{abstract}
With the recent Nobel Prize awarded for radical advances in protein discovery, foundation models (FMs) for exploring large combinatorial spaces promise to revolutionize many scientific fields.
Artificial Life (ALife) has not yet integrated FMs, thus presenting a major opportunity for the field to alleviate the historical burden of relying chiefly on manual design and trial-and-error to discover the configurations of lifelike simulations.
This paper presents, for the first time, a successful realization of this opportunity using vision-language FMs.
The proposed approach, called \textit{Automated Search for Artificial Life} (ASAL),  (1)~finds simulations that produce target phenomena, (2)~discovers simulations that generate temporally open-ended novelty, and (3)~illuminates an entire space of interestingly diverse simulations.
Because of the generality of FMs, ASAL works effectively across a diverse range of ALife substrates including Boids, Particle Life, Game of Life, Lenia, and Neural Cellular Automata.
A major result highlighting the potential of this technique is the discovery of previously unseen Lenia and Boids lifeforms, as well as cellular automata that are open-ended like Conway's Game of Life. 
Additionally, the use of FMs allows for the quantification of previously qualitative phenomena in a human-aligned way.
This new paradigm promises to accelerate ALife research beyond what is possible through human ingenuity alone.
\bigskip

\noindent \textbf{Website: } \href{https://asal.sakana.ai/}{https://asal.sakana.ai/}

\noindent \textbf{Code: }\href{https://github.com/SakanaAI/asal}{https://github.com/SakanaAI/asal}

\end{abstract}

\section{Introduction}

A core philosophy driving Artificial Life (ALife) is to study not only ``life as we know it'' but also ``life as it could be''~\citep{langton1992artificial}.
Because ALife primarily studies life through computational simulations, this approach necessarily means searching through and mapping out an entire \textit{space of possible simulations} rather than investigating any single simulation.
By doing so, researchers can study why and how different simulation configurations give rise to distinct emergent behaviors.
In this paper, we aim, for the first time, to automate this search through simulations with help from foundation models from AI.

While the specific mechanisms for evolution and learning within ALife simulations are rich and diverse, a major obstacle so far to fundamental advances in the field has been the lack of a systematic method for searching through all the possible simulation configurations themselves.
Without such a method, researchers must resort to intuitions and hunches when devising perhaps the most important aspect of an artificial world--the rules of the world itself.

Part of the challenge is that large-scale interactions of simple parts can lead to complex emergent phenomena that are difficult, if not impossible, to predict in advance~\citep{anderson1972more, wolfram2003new}.
This disconnect between the simulation configuration and its resulting behavior makes it difficult to intuitively design simulations that exhibit self-replication, ecosystem-like dynamics, or open-ended properties.
As a result, the field often delivers manually designed simulations tailored to simple and anticipated outcomes, limiting the potential for unexpected discoveries.

Given this present improvisational state of the field, a method to automate the search for simulations themselves would transform the practice of ALife by significantly scaling the scope of exploration.
Instead of probing for rules and interactions that ``feel right'', researchers could refocus their attention to the higher-level question of how to best describe the phenomena we ultimately want to emerge as an outcome, and let the automated process of searching for those outcomes then take its course.

Describing target phenomena for simulations is challenging in its own right, which in part explains why automated search for the right simulation to obtain target phenomena has languished~\citep{stepney2024open, stanley2015greatness}.
Of course, there have been many previous attempts to quantify ALife through intricate measures of life~\citep{sharma2023assembly}, complexity~\citep{mota2013sophistication, lloyd2001measures}, or ``interestingness''~\citep{schmidhuber1997interesting, pathak2017curiosity, secretan2011picbreeder}.
However, these metrics almost always fail to fully capture the nuanced human notions they try to measure~\citep{lloyd2001measures, karwowski2023goodhart}.

While we don't yet understand why or how our universe came to be so complex, rich, and interesting, we can still use it as a guide to create compelling ALife worlds.
Foundation models (FMs) trained on large amounts of natural data possess representations often similar to humans \citep{zhang2018unreasonable, fu2023dreamsim, zhang2023omni} and may even be converging toward a `platonic' representation of the statistics of our real world~\citep{huh2024platonic}.
This novel property makes them appealing candidates for quantifying \textit{human notions of complexity} in ALife.

In this spirit, we propose a new paradigm for ALife research called \textit{Automated Search for Artificial Life} (ASAL).
The researcher starts by defining a set of simulations of interest, referred to as the \textit{substrate}.
Then, as shown in Figure~\ref{fig:teaser}, ASAL enables three distinct methods for FMs to identify interesting ALife simulations:
\begin{enumerate}
    \item \textbf{Supervised Target }
    Searching for a simulation that produces a specified target event or sequence of events, facilitating the discovery of arbitrary worlds or those similar to our own.
    \item \textbf{Open-Endedness }
    Searching for a simulation that produces temporally open-ended novelty in the FM representation space, thereby discovering worlds that are persistently interesting to a \textit{human observer}.
    \item \textbf{Illumination }
    Searching for a set of \textit{interestingly} diverse simulations, enabling the illumination of alien worlds.
\end{enumerate}

The promise of this new automated approach is demonstrated on a diverse range of ALife substrates including Boids, Particle Life, Game of Life, Lenia, and Neural Cellular Automatas.
In each substrate, ASAL discovered previously unseen lifeforms and expanded the frontier of emergent structures in ALife.
For~example, ASAL revealed exotic flocking patterns in Boids, new self-organizing cells in Lenia, and identified life-like cellular automata which are open-ended like the famous Conway's Game of Life.
In addition to facilitating discovery, ASAL's FM framework allows for quantitative analysis of previously qualitative phenomena in ALife simulations, providing a human-like approach to measuring complexity.
ASAL is agnostic to both the specific FM and the simulation substrate, enabling compatibility with future FMs and ALife substrates.

Overall, our new FM-based paradigm serves as a valuable tool for future ALife research by stepping towards the field's ultimate goal of exploring the vast space of artificial life forms.
To the best of our knowledge, this is the first work to drive ALife simulation discovery through foundation models.

\section{Related Works}
\paragraph{ALife Motivations}
ALife is a diverse field that studies life through artificial simulations, with the key difference from biology being its pursuit of the general properties of all life rather than just the specific instantiation of life on Earth~\citep{langton1992artificial}.
ALife systems range widely from cellular automata to neural network agents, but the field generally focuses on emergent phenomena like self-organization, open-ended evolution, agentic behavior, and collective intelligence~\citep{aguilar2014past}.
ALife studies these phenomena to understand of the fundamental principles that govern all life-like systems, and potentially use this information to recreate life in an artificial setting.
Ideas from ALife have also trickled into AI~\citep{nisioti2024text, risi2021future, jiang2023general, ha2022collective, clune2019ai, hughes2024open, schmidhuber2013powerplay, silver2017mastering, wang2020enhanced, dennis2020emergent, parker2022evolving}.

\paragraph{ALife Substrates}
Many substrates are used in ALife to study phenomena at different levels of abstraction and imposed structure, ranging from modeling chemistry to societies.
Conway's Game of Life and other ``Life-Like'' cellular automatas (CA) were critical to the field in the early days and are used to study how complexity may emerge from simple local rules~\citep{games1970fantastic, wolfram2003new}.
Lenia generalizes these to continuous dynamics \citep{chan2018lenia, chan2020lenia, chan2023towards}, and inspired future variants like FlowLenia~\citep{plantec2023flow} and ParticleLenia~\citep{mordvintsev2022particle}.
Neural Cellular Automata (NCA) further generalize Lenia by modeling any continuous CA by parameterizing the update rule with a neural network \citep{mordvintsev2020growing}. 
Instead of operating in a 2-D grid, ParticleLife (or Clusters) uses particles in euclidean space interacting with each other to create dynamic self-organizing patterns \citep{ventrella2017clusters, mohr2023particlelife}.
Similarly, Boids uses bird-like objects to model the flocking behavior of real birds and fish \citep{reynolds1987flocks}.
BioMaker CA and JaxLife are structured substrates designed to study the agentic behavior of plants and societies \citep{lu2024jaxlife, randazzo2023biomaker}, joining other notable substrates like Evolved Virtual Creatures~\citep{SimsKarl1994}, Polyworld~\citep{yaeger1994computational}, and Neural MMO~\citep{suarez2019neural} that focus on studying survival in natural environments.
Some substrates are constructed to study more exotic phenomena, such as self-replicating programs~\citep{ofria2004avida, alakuijala2024computational}.
All of these systems are designed to explore specific aspects of life, with a common theme of emergence from simple components.

Our method aims to be substrate agnostic, with the constraint that the substrate can be displayed as an image.
The majority of ALife substrates are made renderable for human interpretability, including the non-visual substrates like the program based ones \citep{alakuijala2024computational}.

\paragraph{Automatic Search Algorithms in ALife}
Automatic search has been a useful tool in ALife whenever the target outcome is well defined.
In the early days, genetic algorithms were used to evolve CAs to produce a target computation~\citep{packard1988adaptation, mitchell1993revisiting}.
More recently, \citet{chan2020lenia} uses objectives specific to Lenia (e.g.\ speed) to search for new organisms.
BioMaker uses a objective which measured how many agents survived after many timesteps~\citep{randazzo2023biomaker}.
NCA's objective is to make the final state look like a given target image~\citep{mordvintsev2020growing}.

Novelty search~\citep{lehman2011abandoning, lehman2011evolving} is a search algorithm inspired by ALife but requires a good representation space to be effective.
MAP-Elites~\citep{mouret2015illuminating} is a search algorithm which searches along two predefined axis of interest.

Intrinsically motivated discovery uses search in the representation space of an autoencoder to discover new self-organizing patterns \citep{reinke2019intrinsically, falk2024curiosity}.
LeniaBreeder~\citep{faldor2024toward} uses MAP-Elites~\citep{mouret2015illuminating} to search for organisms which have specific properties for e.g.\ mass and speed.
Although LeniaBreeder does provide an unsupervised algorithm as well, it has only been shown to work in Lenia and cannot guide searches via prompt or an open-ended formulation.
Additionally, learning an autoencoder using only images from the substrate may not learn human-aligned representations due to the lack of data diversity~\citep{friedman2022vendi}.

\paragraph{Characterizing Emergence}
Many attempt have been made to quantify complexity~\citep{lloyd2001measures, mitchell2009complexity}.
In information theory, Kolmogorov complexity measures the length of the shortest computer program that produces an artifact~\citep{kolmogorov1998tables}.
Rather than measuring the complexity of an artifact directly, sophistication measures the complexity of the set in which the artifact is a ``generic'' member~\citep{mota2013sophistication}.
Stemming from biochemistry, assembly theory hopes to quantify evolution by measuring the minimal number of steps required to assemble an artifact from atomic building blocks or previously assembled pieces~\citep{sharma2023assembly}.
Although theoretically compelling, these metrics are not computable or fail to capture the nuanced human notions of complexity~\citep{lloyd2001measures, karwowski2023goodhart}.

\citet{wolfram2003new} claims that most complex systems are subject to computational irreducibility, meaning that the emergent behavior of such systems cannot be reduced to a simple theory.

Open-endedness (OE) is the ability of a system to keep generating interesting artifacts forever, which is one of the defining features of natural evolution~\citep{stanley2017open}.
Many necessary conditions for OE have been identified but are yet to be realized \citep{soros2014identifying}.
There have been some attempts at quantifying OE~\citep{hughes2024open, wang2020enhanced}, but some argue that OE cannot be quantified by definition~\citep{stepney2024open}.
In one case study, human intervention was essential for achieving OE evolution~\citep{secretan2011picbreeder}, suggesting that OE may depend on novelty within a particular representation space.
This aligns with the idea that while all interesting things are novel, not all novel things are inherently interesting~\citep{stanley2015greatness, schmidhuber1997interesting, herrmann2022learning}.

\paragraph{Foundation Models for Automatic Search}

Large pretrained neural networks, often referred to as foundation models (FMs)~\citep{bommasani2021opportunities, kaplan2020scaling}, are currently revolutionizing many scientific domains.
In medicine, FMs transformed the drug discovery process by enabling accurate predictions of protein folding~\citep{jumper2021highly}.
In robotics, LLMs have automated the design of reward functions, alleviating a typically tedious task for humans~\citep{ma2023eureka}.
In physics, large models are used to predict complex systems and are later distilled into symbolic equations~\citep{cranmer2020discovering}.
AI systems have even reached Olympiad-level performance in solving geometry problems~\citep{trinh2024solving}.

The potential of using FMs, particularly large language models (LLMs), for ALife and vice versa was highlighted in~\citet{nisioti2024text}.
LLMs have been applied in ALife-like contexts as code mutation operators \citep{lehman2023evolution, faldor2024omni} and for proposing next goals \citep{zhang2023omni, todd2023level}.
These applications are limited to text-based search spaces and often rely on LLMs within the inner simulation loop, which is not applicable when analyzing dynamical systems governed by simple update rules.

In this work, we use CLIP, an image-language embedding model trained with contrastive learning to align text-image representations on an internet-scale dataset~\citep{radford2021learning}.
CLIP’s simplicity and generality enable it to effectively render images from vector strokes~\citep{frans2022clipdraw} or shapes~\citep{tian2022modern}, and to guide generation in other generative models such as VQ-GAN~\citep{crowson2022vqgan} and Diffusion~\citep{kim2022diffusionclip}.
We apply CLIP to search for ALife simulations instead of static images, resulting in analyzable artifacts that provide valuable insights for ALife research.

\begin{figure}[!h]
    \centering
    \includegraphics[width=\textwidth]{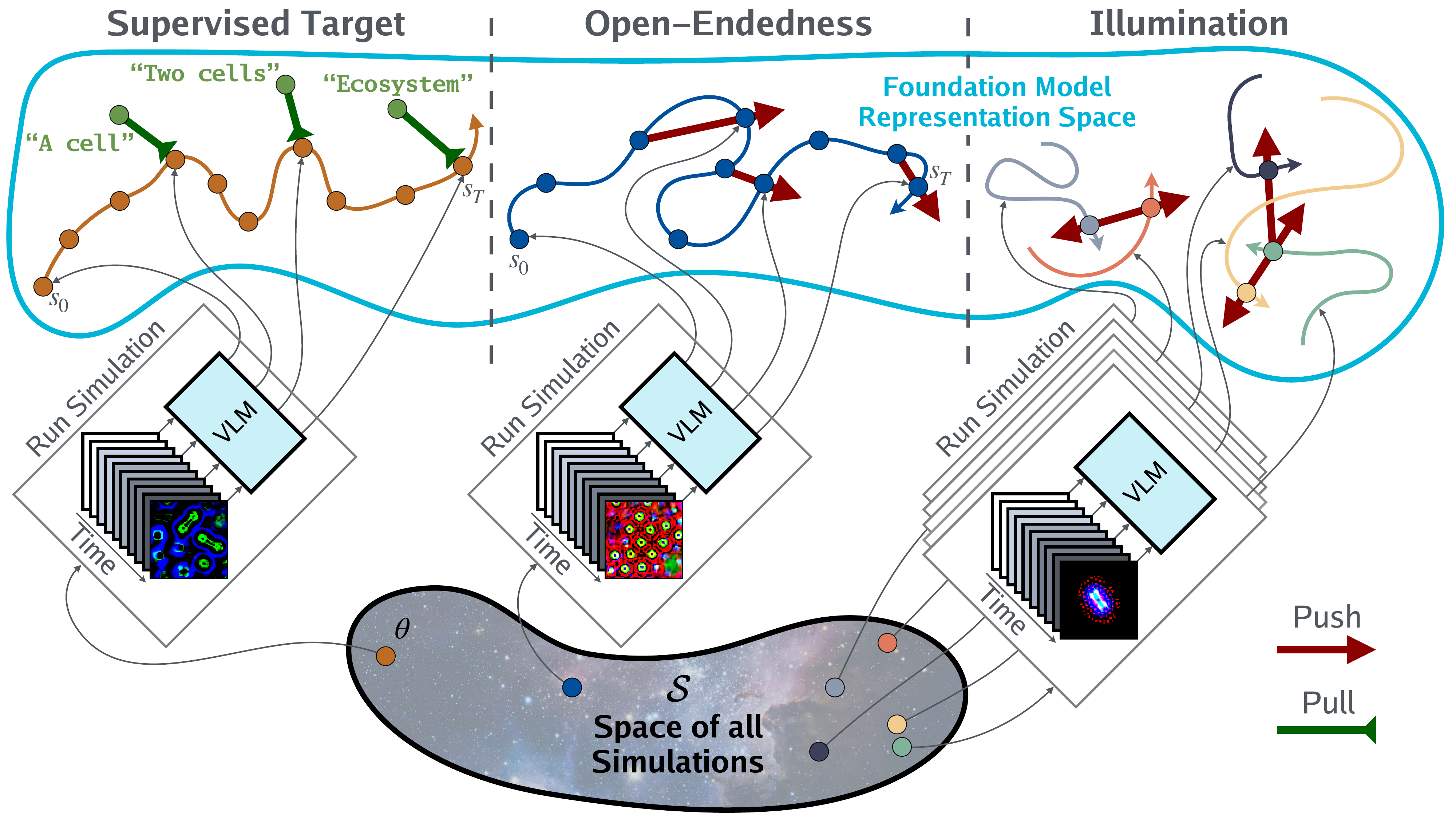}
    \caption{
    \textbf{ASAL:}
    Our proposed framework, ASAL, uses vision-language foundation models to discover ALife simulations by formulating the processes as three search problems.
    \textbf{Supervised Target:}
    To find target simulations, ASAL searches for a simulation which produces a trajectory in the foundation model space that aligns with a given sequence of prompts.
    \textbf{Open-Endedness:}
    To find open-ended simulations, ASAL searches for a simulation which produces a trajectory that has high historical novelty during each timestep.
    \textbf{Illumination:}
    To illuminate the set of simulations, ASAL searches for a set of diverse simulations which are far from their nearest neighbor.
    }
    \label{fig:method_overview}
\end{figure}

\section{Methods: Automated Search for Artificial Life}

Figure~\ref{fig:method_overview} depicts our proposed paradigm, \textit{Automated Search for Artificial Life} (ASAL), which includes three algorithms built on vision-language FMs.
Each method discovers ALife simulations through a different kind of automated search.
Before diving into the details, relevant concepts and notations are introduced next.

An ALife substrate, $\mathcal{S}$ encompasses any set of ALife simulations of interest (e.g.\ the set of all Lenia simulations).
These could vary in the initial states, transition rules, or both.
$\mathcal{S}$ is parameterized by $\theta$, which defines a single simulation with three components:
\begin{itemize}
    \item the initial state distribution, $\texttt{Init}_\theta$,
    \item the forward dynamics step function, $\texttt{Step}_\theta$,
    \item the rendering function, $\texttt{Render}_\theta$, which transforms the state into an image.
\end{itemize}
While parameterizing and searching for a renderer is often not needed, it becomes necessary when dealing with state values that are uninterpretable a priori.
This often happens in less structured substrates, like Neural Cellular Automata~\citep{mordvintsev2020growing}, where the state tensor has no semantic meaning to the human.

Chaining these terms together, we define a function of $\theta$ that samples an initial state $s_0$, runs the simulation for $T$ steps, and renders that final state as an image:
\begin{equation}
\texttt{RS}^T(\theta)
=
\texttt{Render}_\theta\left(
\underbrace{
\texttt{Step}_\theta(\texttt{Step}_\theta(\texttt{Step}_\theta(\dots (s_0))))
}_{\text{Simulate } T \text{ timesteps}}
\right)
,\quad \text{where} \quad
s_0\sim\texttt{Init}_\theta
\end{equation}

Finally, two additional functions $\texttt{VLM}_\texttt{img}(\cdot)$ and $\texttt{VLM}_\texttt{txt}(\cdot)$ embed images and natural language text through the vision-language FM, along with a corresponding inner product $\left<\cdot, \cdot\right>$ to facilitate similarity measurements for that embedding space.

\colorlet{color_solution}{Plum}
\colorlet{color_search}{Red}
\colorlet{color_fm}{RoyalBlue}
\colorlet{color_img}{Brown}
\colorlet{color_txt}{ForestGreen}
\colorlet{color_novelty}{ForestGreen}

\subsection{Supervised Target}
An important goal in ALife is to find simulations where a desired event or sequence of events take place~\citep{bedau2000open}.
Such discovery would allow researchers to identify worlds similar to our own or test whether certain counterfactual evolutionary trajectories are even possible in the given substrate, thus giving insights about the feasibility of certain lifeforms.

For this purpose, ASAL \textcolor{color_search}{\textbf{searches}} for \textcolor{color_solution}{\textbf{a simulation}} that \textcolor{color_img}{\textbf{produces images}} that match a \textcolor{color_txt}{\textbf{target natural language prompt}} in the \textcolor{color_fm}{\textbf{FM's representation}} (colors match the equation below).
The researcher has control of which prompt, if any, to apply at every simulation timestep, $T$, thus providing temporal guidance.

\begin{equation}
\textcolor{color_solution}{\theta^*}
=
\textcolor{color_search}{\argmax_\theta}
\mathbb{E}_{T}
\left[
\left<
\textcolor{color_fm}{\texttt{VLM}_\texttt{img}}\left(
\textcolor{color_img}{\texttt{RS}^{T}(\theta)}
\right)
,
\textcolor{color_fm}{\texttt{VLM}_\texttt{txt}}\left(
\textcolor{color_txt}{\texttt{prompt}_T}
\right)
\right>
\right]
\label{eq:supervised}
\end{equation}

This equation uses an $\argmax$ because it is \textit{maximizing} the alignment between the simulation image and the prompt.
The expectation over $T$ is over timesteps the user has prompts specified for.

\subsection{Open-Endedness}
A grand challenge of ALife is finding open-ended simulations \citep{stanley2017open, bedau2000open}.
Finding such worlds is necessary for replicating the explosion of never-ending interesting novelty that the real world is known for.

Although open-endedness is subjective and hard to define, novelty in the right representation space captures a general notion of open-endedness \citep{lehman2011abandoning, stanley2017open, hughes2024open}.
This formulation outsources the subjectivity of measuring open-endedness to the construction of the representation function, which embodies the observer.
In this paper, the vision-language FM representations act as a proxy for a human's representation~\citep{zhang2018unreasonable}.

With this novel capability, ASAL \textcolor{color_search}{\textbf{searches}} for \textcolor{color_solution}{\textbf{a simulation}} which \textcolor{color_img}{\textbf{produces images}} that are \textcolor{color_novelty}{\textbf{historically novel}} in the \textcolor{color_fm}{\textbf{FM's representation}}.
Some preliminary experiments showed that historical nearest neighbor novelty produces better results than variance based novelty.

\begin{equation}
\textcolor{color_solution}{\theta^*}
=
\textcolor{color_search}{\argmin_\theta}
\mathbb{E}_{T}
\left[
\textcolor{color_novelty}{\max_{T'<T}}
\left<
\textcolor{color_fm}{\texttt{VLM}_\texttt{img}}\left(
\textcolor{color_img}{\texttt{RS}^{T}(\theta)}
\right)
,
\textcolor{color_fm}{\texttt{VLM}_\texttt{img}}\left(
\textcolor{color_novelty}{\texttt{RS}^{T'}(\theta)}
\right)
\right>
\right]
\label{eq:oe}
\end{equation}

This equation uses an $\argmin$ because it is \textit{minimizing} the alignment between the simulation image at timestep $T$ and its historically closest neighbor at $T'$.
The expected value term acts as an ``open-endedness'' score for a given simulation.

\subsection{Illumination}
Another key goal in ALife is to automatically illuminate the entire space of diverse phonemena that can emerge within a substrate, motivated by the quest to understand ``life as it could be''~\citep{bedau2000open}.
Such illumination is the first step to mapping out and taxonomizing an entire substrate.

Towards this aim, ASAL \textcolor{color_search}{\textbf{searches}} for a \textcolor{color_solution}{\textbf{set of simulations}} that \textcolor{color_img}{\textbf{produce images}} that are far from their \textcolor{color_novelty}{\textbf{nearest neighbor}} in the \textcolor{color_fm}{\textbf{FM's representation}}.
We find that nearest neighbor diversity produces better illumination than variance based diversity.

\begin{equation}
\textcolor{color_solution}{\{\theta_0^*,\dots,\theta_n^*\}}
=
\textcolor{color_search}{\argmin_{\theta_0,\dots,\theta_n}}
\mathbb{E}_{\theta,T}
\left[
\textcolor{color_novelty}{\max_{\theta'\neq \theta}}
\left<
\textcolor{color_fm}{\texttt{VLM}_\texttt{img}}\left(
\textcolor{color_img}{\texttt{RS}^{T}(\theta)}
\right)
,
\textcolor{color_fm}{\texttt{VLM}_\texttt{img}}\left(
\textcolor{color_novelty}{\texttt{RS}^{T}(\theta')}
\right)
\right>
\right]
\label{eq:illumination}
\end{equation}

This equation uses an $\argmin$ because it is \textit{minimizing} the alignment between simulation $\theta$ and its nearest other simulation $\theta'$.
The expected value term acts as an ``diversity'' score for a given set of simulations.

\section{Experiments}

This section experimentally validates the effectiveness of ASAL across various substrates, then presents novel quantitative analyses of some of the discovered simulations, facilitated by the FM.
Before presenting the experiments, here is a summary of the FMs and substrates used.
The appendix includes additional details about the substrates and experimental setups, as well as supplementary experiments.

\paragraph{Foundation Models}
The following vision foundation models were chosen for their generality.

\begin{itemize}
    \item
\textbf{CLIP} (Contrastive Language-Image Pretraining) is a vision-language FM that learns general-purpose image and text representations by aligning their latent spaces through contrastive pretraining on an internet scale dataset~\citep{radford2021learning}.
CLIP explicitly provides $\texttt{VLM}_\texttt{img}(\cdot)$ and $\texttt{VLM}_\texttt{txt}(\cdot)$.

    \item
\textbf{DINOv2} (\textbf{Di}stillation with \textbf{No} Labels) is a vision-only FM that learns visual representations by using a self-supervised teacher-student framework on a large image dataset~\citep{oquab2023dinov2}.
DINOv2 only provides $\texttt{VLM}_\texttt{img}(\cdot)$, and so cannot be used with ASAL's supervised target search.
\end{itemize}

\paragraph{Substrates}
The following substrates were chosen for their diverse mechanisms and strong visual elements, making them well-suited for vision foundation models.
\begin{itemize}
    \item
\textbf{Boids} simulates $N$ ``bird-like objects'' (boids) moving in a 2-D Euclidean space~\citep{reynolds1987flocks}.
All boids weight-share a single neural network that steers each boid left or right based on the $K$ nearby boids in its local frame of reference .
The substrate is the weight-space of the neural network.
    \item
\textbf{Particle Life} (or Clusters) simulates $N$ particles, each belonging to one of $K$ types, interacting in a 2-D Euclidean space \citep{ventrella2017clusters, mohr2023particlelife}.
The substrate is the space of $K\times K$ interaction matrices and the $\beta$ parameters to determine how close particles can get to each other.
The initial state is sampled randomly and the particles self-organize to form dynamic patterns.
    \item
\textbf{Life-like Cellular Automata} (CA) generalizes Conway’s Game of Life to all binary-state CA that operate in 2-D lattice where the state transition depends solely on the number of living Moore neighbors and the current state of the cell \citep{wojtowicz2024cellular}. 
The substrate has $2^{18}=262$,$144$ possible simulations.
    \item
\textbf{Lenia} generalizes Conway’s Game of Life to continuous space and time, allowing for higher dimensions, multiple kernels, and multiple channels~\citep{chan2018lenia}.
We utilize the LeniaBreeder codebase, which defines the substrate with $45$ dimensions for dynamics and $32 \times 32 \times 3 = 3$,$072$ dimensions for the initial state \citep{faldor2024toward}.
The search space is centered at a solution found by~\citet{chan2020lenia}.
    \item
\textbf{Neural Cellular Automata} (NCA) parameterizes any continuous cellular automata by representing the local transition function through a neural network \citep{mordvintsev2020growing}.
The substrate is the weight-space of the neural network.
\end{itemize}

\begin{figure}[h]
  \centering
  \includegraphics[width=1.0\linewidth]{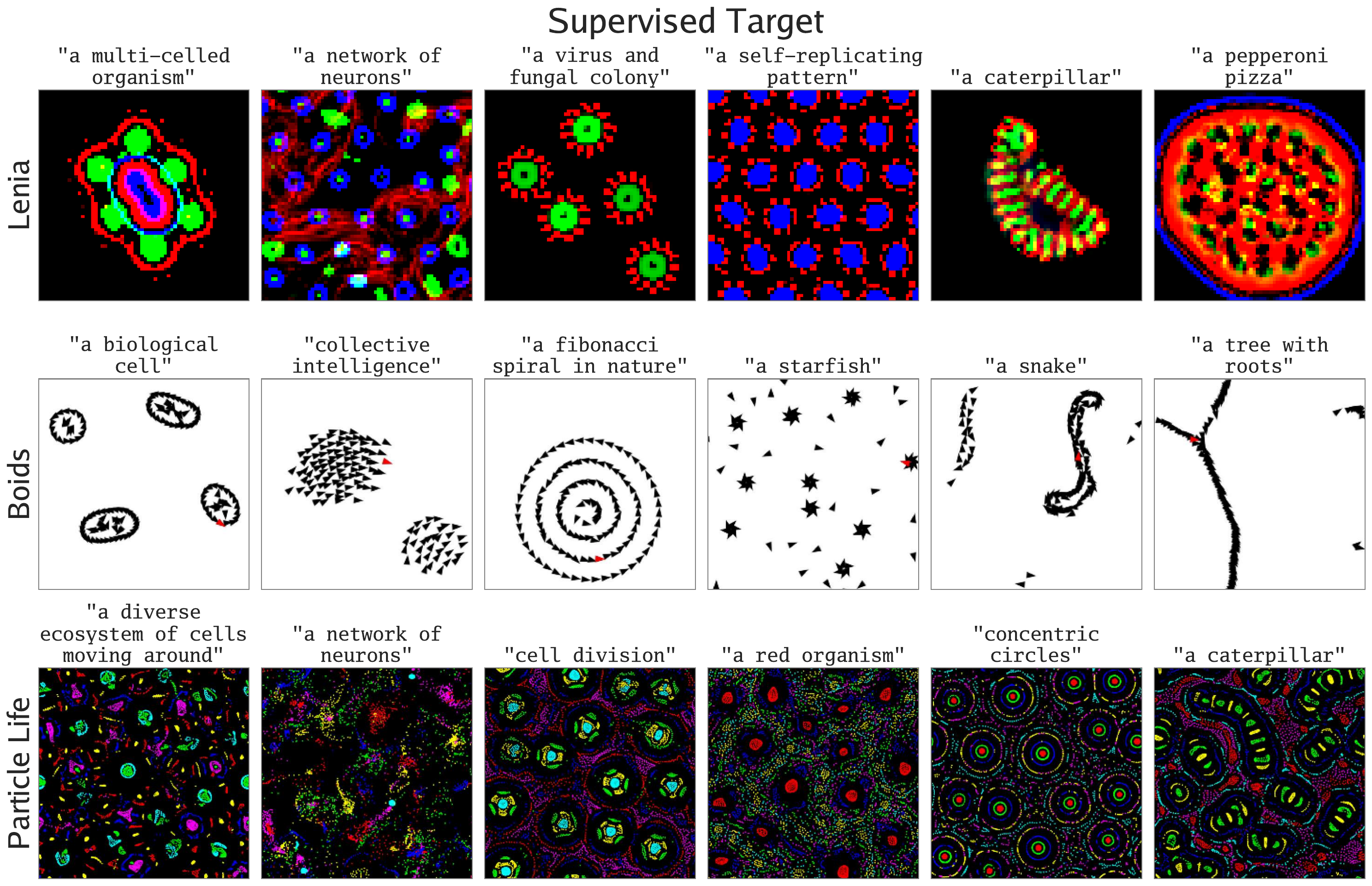}
  \caption{
    \textbf{Discovered target simulations:}
    Using Equation~\ref{eq:supervised}, ASAL discovered simulations that result in a final state which matches the specified prompt.
    Results are shown for three different substrates.
    More prompts can be seen in appendix Figures~\ref{fig:supervised_extra_lenia}, \ref{fig:supervised_extra_boids}, and \ref{fig:supervised_extra_plife}.
  }
  \label{fig:supervised}
\end{figure}

\begin{figure}[htp]
    \centering
    \includegraphics[width=\textwidth]{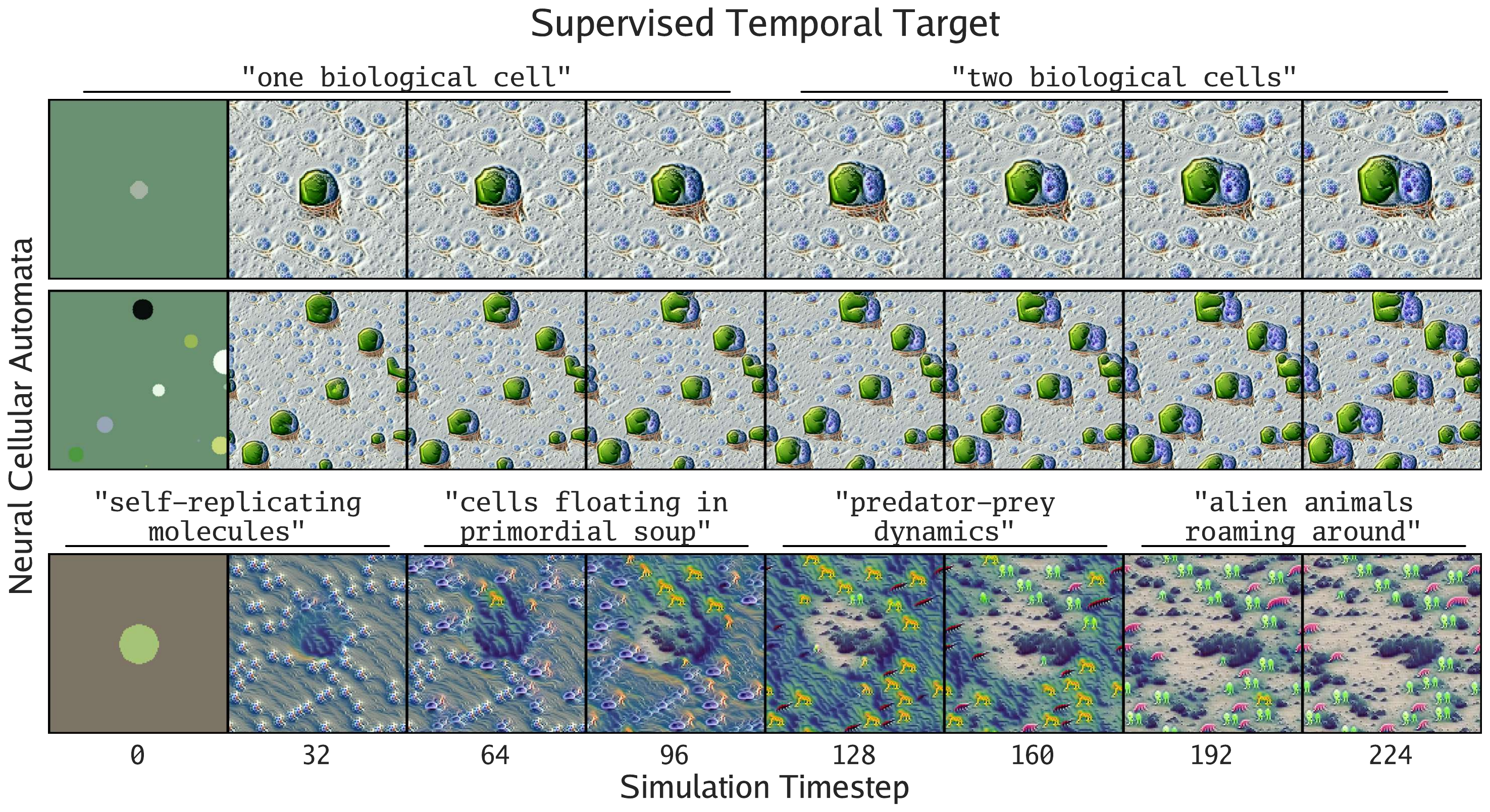}
    \caption{
    \textbf{Discovered temporal target simulations: }
    Using Equation~\ref{eq:supervised}, ASAL discovered simulations that produce a sequence of events which match a list of prompts.
    The second row shows how the first simulation generalizes to a different initial state.
    The results are shown for the NCA substrate.
  }
  \label{fig:nca_temporal}
\end{figure}

\subsection{Searching for Target Simulations}
This section explores both single targets and sequences of targets over time.

\subsubsection{Single Target}
The effectiveness of searching for target simulations specified by a single prompt is explored in Lenia, Boids, and Particle Life.
Equation~\ref{eq:supervised} is optimized with the prompt applied once after $T$ simulation timesteps.
CLIP is the FM and Sep-CMA-ES \citep{ros2008simple} is the optimization algorithm.

Figure~\ref{fig:supervised} shows the optimization works well from a qualitative perspective at finding simulations matching the specified prompt.
The results of more prompts are shown in appendix Figure~\ref{fig:supervised_extra_lenia}, Figure~\ref{fig:supervised_extra_boids} and Figure~\ref{fig:supervised_extra_plife}.
Some of the failure modes suggest that when optimization fails, it is often caused by the lack of expressivity of the substrate rather than the optimization process itself.

\subsubsection{Temporal Targets}
We investigate the effectiveness of searching for simulations producing a target sequence of events using the NCA substrate, because it is the most expressive substrate.
We optimize Equation~\ref{eq:supervised} with a list of prompts, each applied at evenly spaced time intervals of the simulation rollout.
We use CLIP for the FM.
Following the original NCA paper, we use backpropagation through time and gradient descent with the Adam optimizer for the optimization algorithm \citep{mordvintsev2020growing}.

Figure~\ref{fig:nca_temporal} shows it is possible to find simulations that produce trajectories following a sequence of prompts.
By specifying the desired evolutionary trajectories and employing a constraining substrate, ASAL can identify update rules that embody the essence of the desired evolutionary process.
For instance, when the sequence of prompts is ``\texttt{one cell}'' then ``\texttt{two cells}'', the corresponding update rule inherently enables self-replication.

\begin{figure}[htbp]
    \centering
    \begin{subfigure}{\textwidth}
        \centering
        \includegraphics[width=1.0\textwidth]{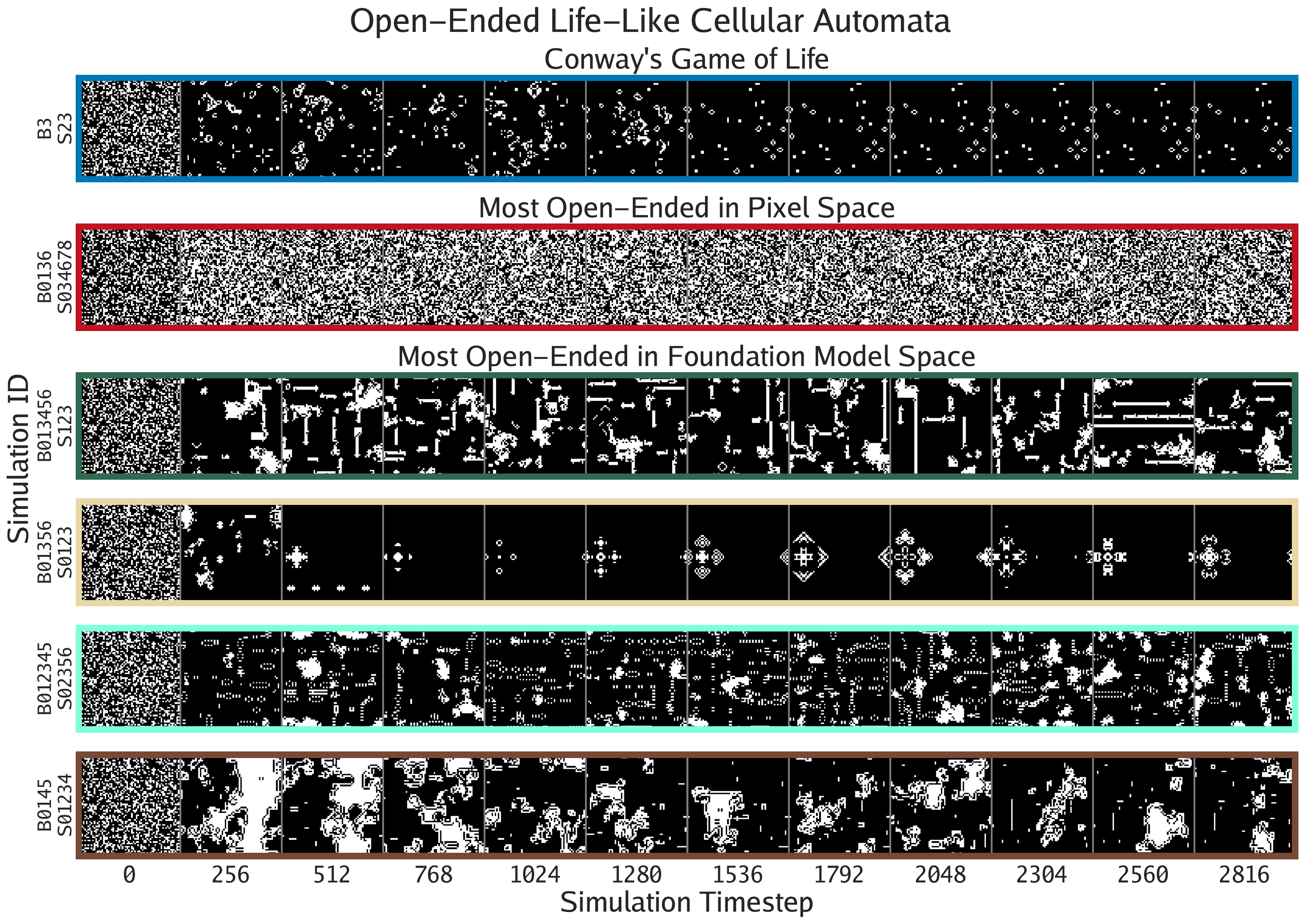}
        \caption{}
        \label{fig:oe_gol_1}
    \end{subfigure}
    
    \begin{subfigure}{0.48\textwidth}
        \centering
        \includegraphics[width=\textwidth]{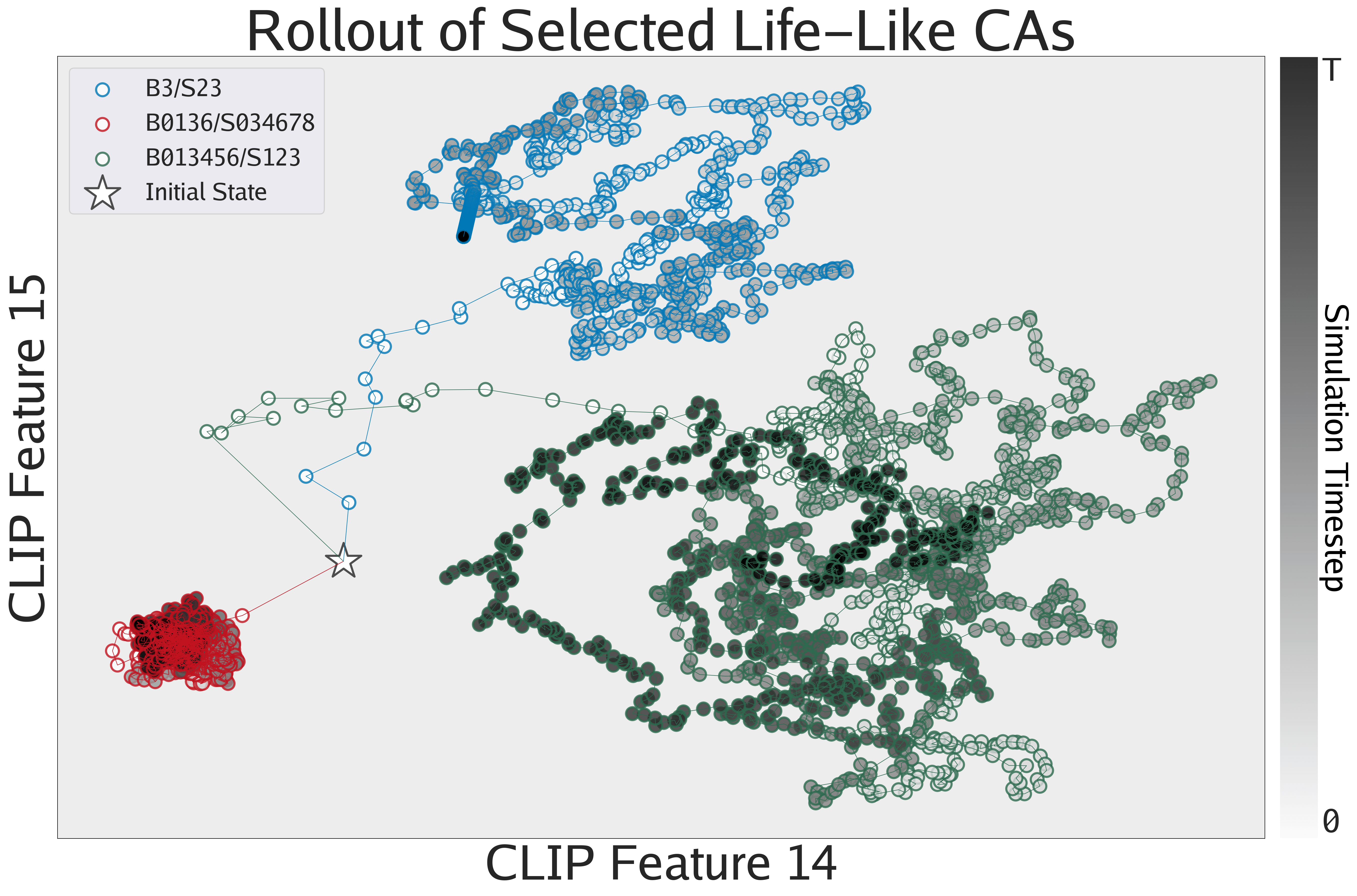}
        \caption{}
        \label{fig:oe_gol_2}
    \end{subfigure}
    \hfill
    \begin{subfigure}{0.49\textwidth}
        \centering
        \includegraphics[width=\textwidth]{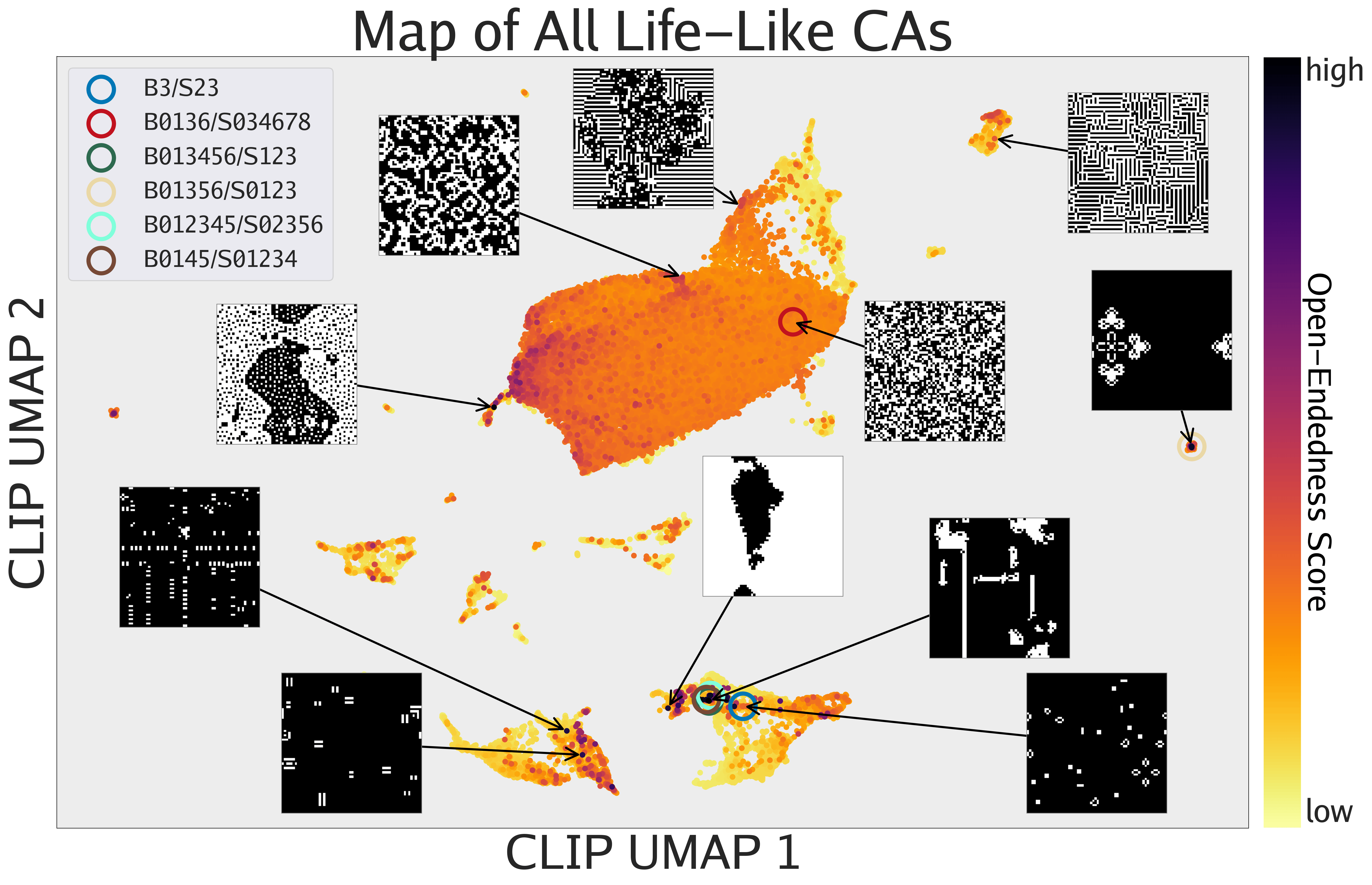}
        \caption{}
        \label{fig:oe_gol_3}
    \end{subfigure}
    \caption{
        \textbf{Discovered open-ended simulations: }
        Using Equation~\ref{eq:oe}, ASAL discovered open-ended simulations in the Life-Like CAs substrate.
        Simulations are labeled in Golly notation~\citep{eppstein2010growth} to denote the number of living neighbors required for birth and survival.
        \textbf{(a)}
        The discovered CAs rendered over a simulation rollout.
        \textbf{(b)}
        The temporal trajectories of three simulations in CLIP space.
        The pixel-space simulation (red) exhibits a convergent trajectory, whereas the FM-space simulation (green) demonstrates a more divergent trajectory, even exceeding that of Conway's Game of Life (blue).
        \textbf{(c)}
        All Life-like CAs plotted based on the UMAP~\citep{mcinnes2018umap} projection of the CLIP embedding of their final state, colored by open-endedness score.
        The resulting structure reveals distinct islands of similar simulations, with the most open-ended CAs grouped together near the bottom.
    }
    \label{fig:oe_gol}
\end{figure}

\subsection{Searching for Open-Ended Simulations}
To investigate the effectiveness of searching for open-ended simulations, we use the Life-Like CAs substrate and optimize the open-ended score from Equation~\ref{eq:oe}.
CLIP serves as the FM.
Because the search space is relatively small with only $262$,$144$ simulations, brute force search is employed.
We do not search for open-endedness in the Lenia or Boids substrates because preliminary experiments hinted it is hard to sustain novelty in these substrates.
Similarly, we omit the NCA substrate, which we suspect is overly expressive and may fail to yield meaningful structure---though it remains a promising direction for future work.

Figure~\ref{fig:oe_gol} reveals the potential for open-endedness in the Life-like CAs.
The famous Conway's Game of Life ranks among the top 5\% most open-ended CAs according to our open-endedness metric from Equation~\ref{eq:oe}.
Figure~\ref{fig:oe_gol_1} shows the most open-ended CAs demonstrate nontrivial dynamic patterns that lie on edge of chaos, since they neither plateau or explode \citep{wolfram2003new}.
Figure~\ref{fig:oe_gol_2} traces out the trajectories of three CAs in CLIP space over simulation time.
Because the FM's representations are related to human representations, producing novelty in the trajectory through the FM's representation space yields a sequence of novelty to a human observer as well.
Figure~\ref{fig:oe_gol_3} visualizes all Life-Like CAs with a UMAP plot of their CLIP embeddings colored by open-endedness score, and shows that meaningful structure emerges: the most open-ended CAs lie close together on a small island outside the main island of simulations.
Additional open-ended CAs discovered are shown in Appendix Figure~\ref{fig:oe_gol_extra}.
Appendix Figure~\ref{fig:atlas_gol} provides a more detailed visualization of the simulations shown in Figure~\ref{fig:oe_gol_3}.

We plot Figure~\ref{fig:oe_gol_2} along two axes of the CLIP space, rather than in UMAP space, because we found that UMAP often distorts temporal trajectories, creating too much structure not actually present in the data.

\begin{figure}[htbp]
    \centering
    \includegraphics[width=\textwidth]{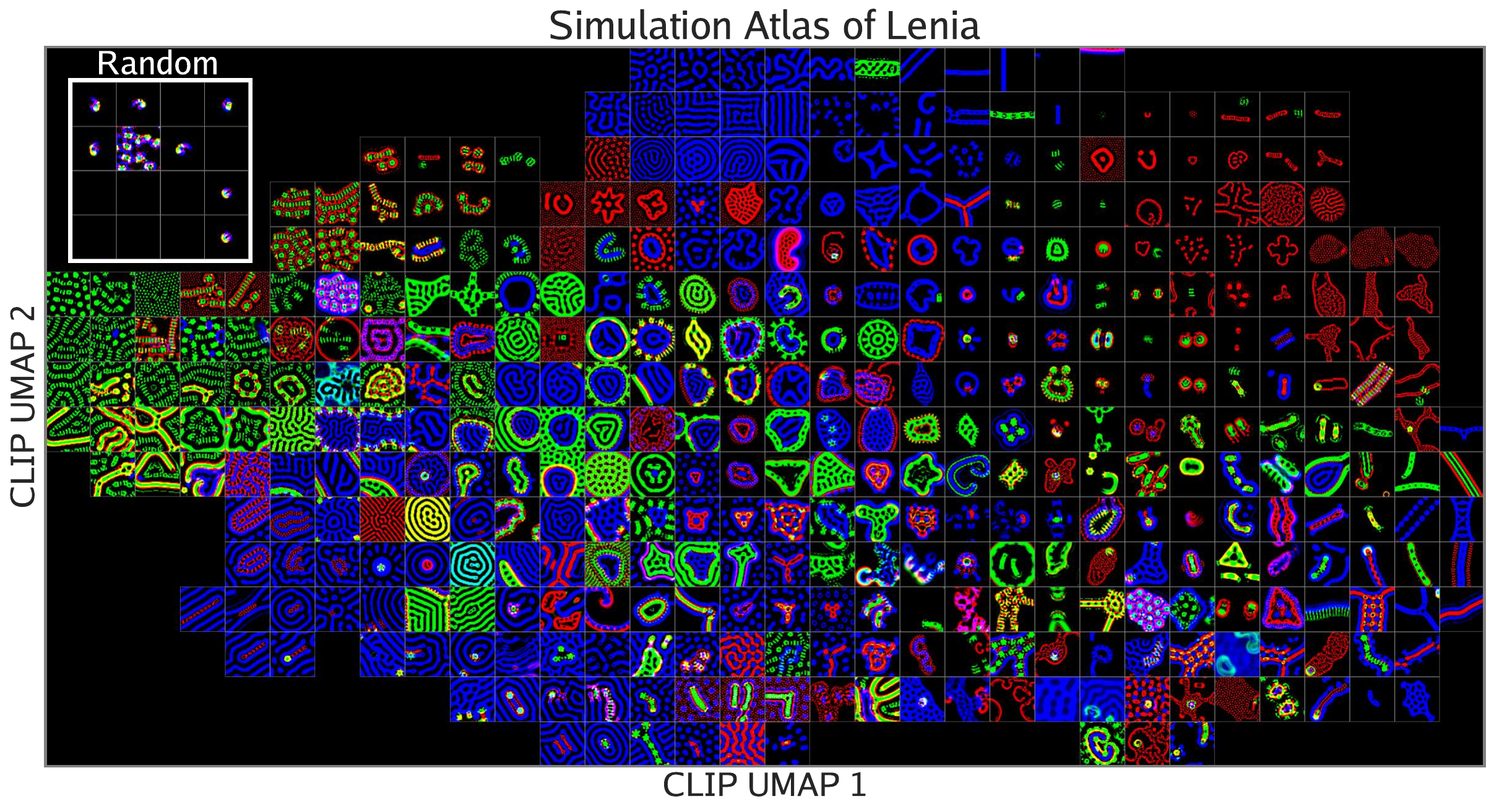}
    \includegraphics[width=\textwidth]{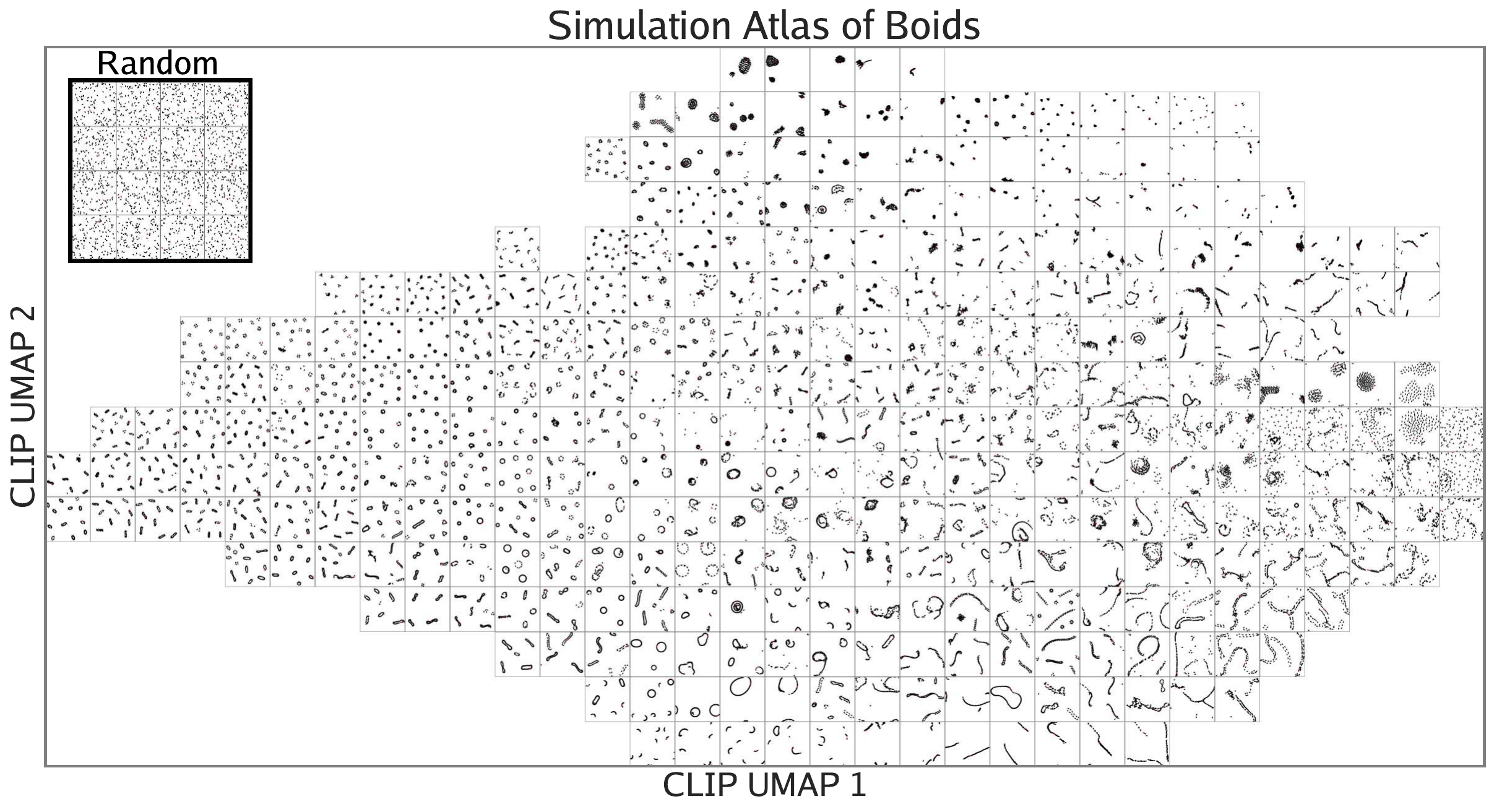}
    \caption{
    \textbf{Simulation Atlas: }
    ASAL discovered a large set of diverse simulations by using the illumination algorithm from Equation~\ref{eq:illumination} on the Lenia and Boids substrates.
    The resulting final states of these simulations are then embedded using CLIP and projected into 2-D with UMAP~\citep{mcinnes2018umap}.
    This space is then grid sampled and the nearest simulation within that tile is shown.
    Empty tiles signify that no simulations exist in that part of CLIP-UMAP space.
    This simulation atlas maps all discovered simulations in an organized manner.
    The top left insets show randomly sampled simulations without illumination.
    Larger simulation atlases can be found in appendix Figure~\ref{fig:atlas_lenia_large} and Figure~\ref{fig:atlas_boids_large}.
    The illumination of Particle Life is in appendix Figure~\ref{fig:illumination_plife}.
    }
    \label{fig:atlas}
\end{figure}

\subsection{Illuminating Entire Substrates}
We use the Lenia and Boids substrates to study the effectiveness of the illumination algorithm from Equation~\ref{eq:illumination}.
CLIP is the FM.
A custom genetic algorithm performs the search: at each generation, it randomly selects parents, creates mutated children, then keeps the most diverse subset of solutions.

The resultant set of simulations is shown in the ``Simulation Atlas'' in Figure~\ref{fig:atlas}.
This visualization highlights the diversity of the discovered behaviors organized by visual similarity.
In Lenia, ASAL discovers many previously unseen lifeforms resembling microscopy images of cells and bacteria, all spatially organized by color and shape.
In Boids, ASAL rediscovers flocking behavior, as well as additional behaviors such as snaking, grouping, circling, and other variations.
Larger simulation atlases are shown in appendix Figure~\ref{fig:atlas_lenia_large} and Figure~\ref{fig:atlas_boids_large}.

\begin{figure}[htp]
    \centering
    \begin{subfigure}{1.0\textwidth}
        \centering
        \includegraphics[width=\textwidth]{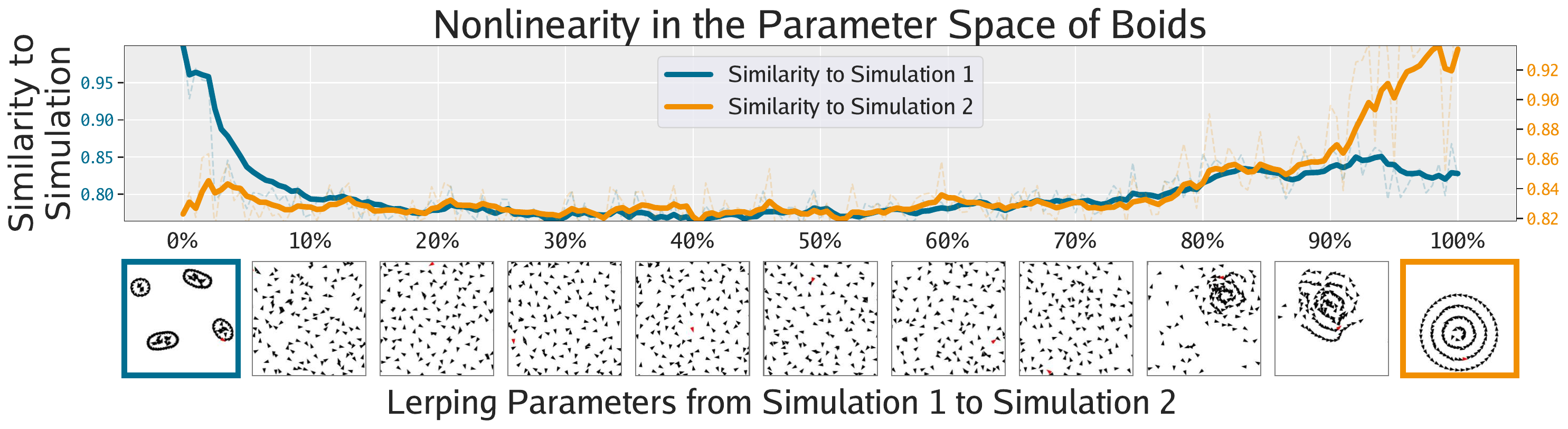}
        \vspace{-6mm}
        \caption{
        The simulation final state as simulation parameters are linearly interpolated from one simulation to another.
        }
        \label{fig:lerping_1}
    \end{subfigure}
    \begin{subfigure}{1.0\textwidth}
        \centering
        \includegraphics[width=\textwidth]{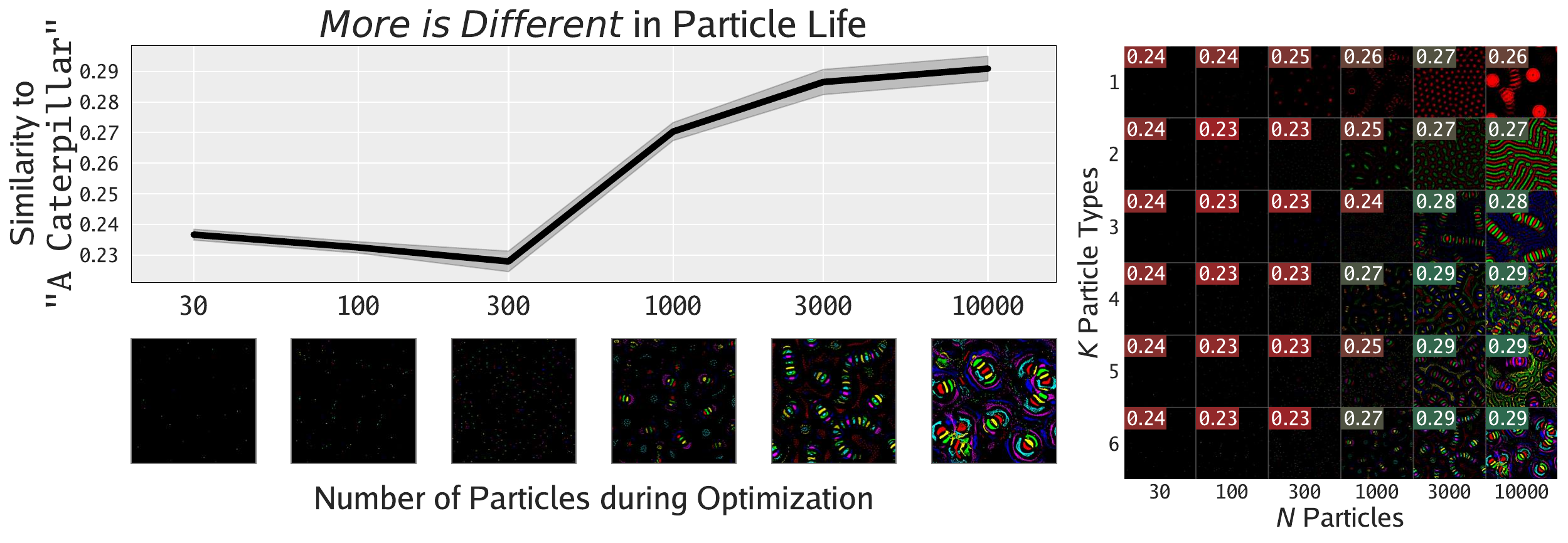}
        \vspace{-6mm}
        \caption{
        Plotting the emergence of ``\texttt{a caterpillar}'' in Particle Life as the number of particles is increased.
        }
        \label{fig:lerping_2}
    \end{subfigure}
    \begin{subfigure}{1.0\textwidth}
        \centering
        \includegraphics[width=\textwidth]{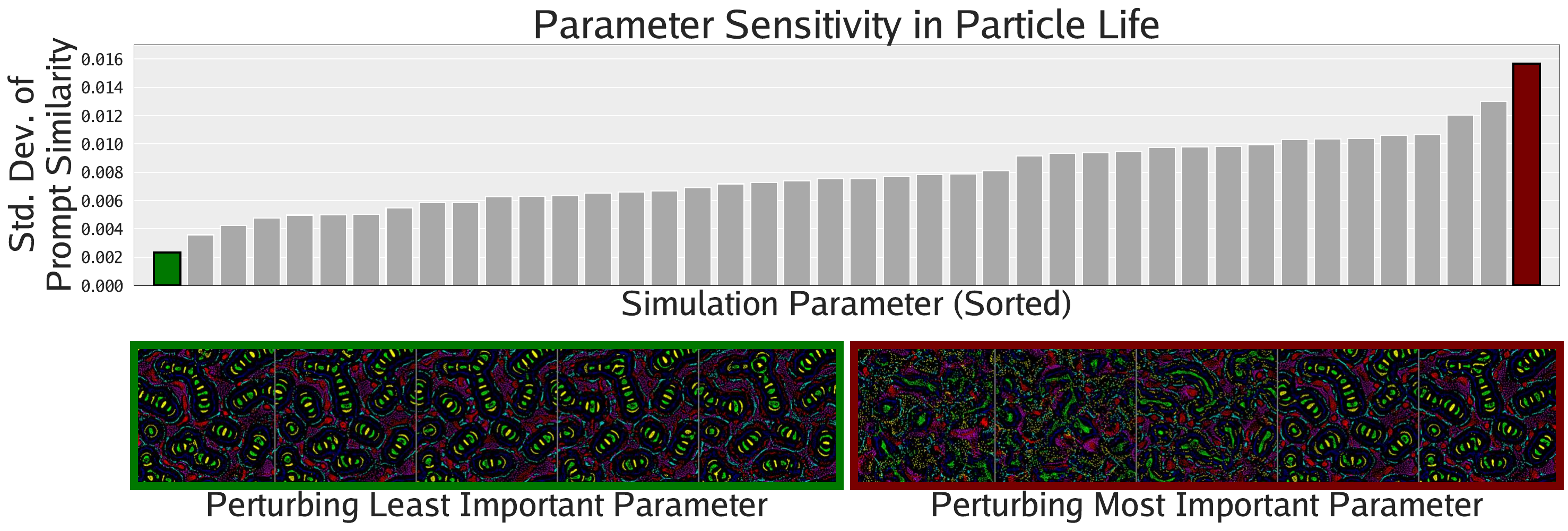}
        \vspace{-6mm}
        \caption{
        Ranking a Particle Life simulation's parameters by importance to the simulation behavior.
        }
        \label{fig:lerping_3}
    \end{subfigure}
    \begin{subfigure}{1.0\textwidth}
        \centering
        \includegraphics[width=\textwidth]{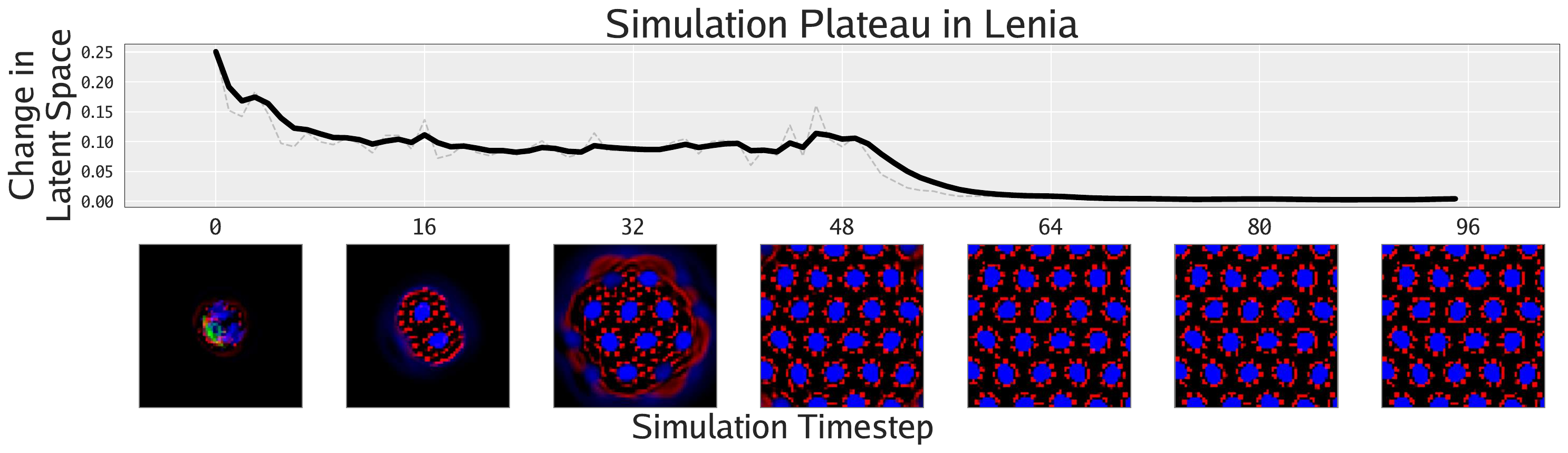}
        \vspace{-6mm}
        \caption{
        Plotting the change of the CLIP embedding over simulation time quantifies the plateau signal in Lenia.
        }
        \label{fig:lerping_4}
    \end{subfigure}
    \caption{
    \textbf{Quantifying ALife:}
    Foundation Models allow human-aligned quantification of previously qualitative phenomena.
    }
    \label{fig:lerping}
\end{figure}

\subsection{Quantifying ALife}
Not only can FMs facilitate the search for interesting phenomena, but they also enable the quantification of phenomena previously only amenable to qualitative analysis, as shown in this section.

Figure~\ref{fig:lerping} shows different ways of quantifying the emergent behaviors of these complex systems.

In Figure~\ref{fig:lerping_1}, we linearly interpolate the parameters between two Boids simulations.
The intermediate simulations lack the characteristics of either simulation and appear disordered, demonstrating the nonlinear, chaotic nature of the boids parameter space.
Importantly, this qualitative observation is now possible to support quantitatively by measuring the CLIP similarity of the final states of the intermediate simulation, $\theta'$, to the original simulation, $\theta$, with $\left<\texttt{VLM}_\texttt{img}\left({\texttt{RS}^{T}(\theta')}\right), \texttt{VLM}_\texttt{img}\left({\texttt{RS}^{T}(\theta)}\right)\right>$.

Figure~\ref{fig:lerping_2} evaluates the effect of the number of particles in Particle Life on its ability to represent certain lifeforms, by measuring the objective in Equation~\ref{eq:supervised}.
In this case we search for ``\texttt{a caterpillar}'' and find that they can only be found when there are at least $1$,$000$ particles in the simulation, matching the ``more is different'' observation \citep{anderson1972more}.

Figure~\ref{fig:lerping_3} quantifies the importance of each of the simulation parameters in Particle Life by individually sweeping each parameter and measuring the resulting standard deviation of the CLIP prompt alignment score, from Equation~\ref{eq:supervised}.
After identifying the most important parameter, this corresponds with the strength of interaction between the green and yellow particles, which is critical for the formation of the caterpillar.

Figure~\ref{fig:lerping_4} shows that the speed of change of the CLIP vector,
$\left\|\frac{\partial}{\partial t} \texttt{VLM}_\texttt{img}\left({\texttt{RS}^{T}(\theta)}\right)\right\|$
, over simulation time for a Lenia simulation.
This metric plateaus exactly when the simulation looks to have become qualitatively static, providing a useful simulation halting condition.

This section showcased four ways of using the FM latent space to measure important aspects of the simulation. 
Additional metrics can be easily constructed by taking an existing mathematical formulation of a phenomenon and replacing the state variables with FM embeddings, or by replacing traditional similarity measures with FM-based similarity.

\begin{figure}[ht]
  \centering
  \includegraphics[width=1.0\linewidth]{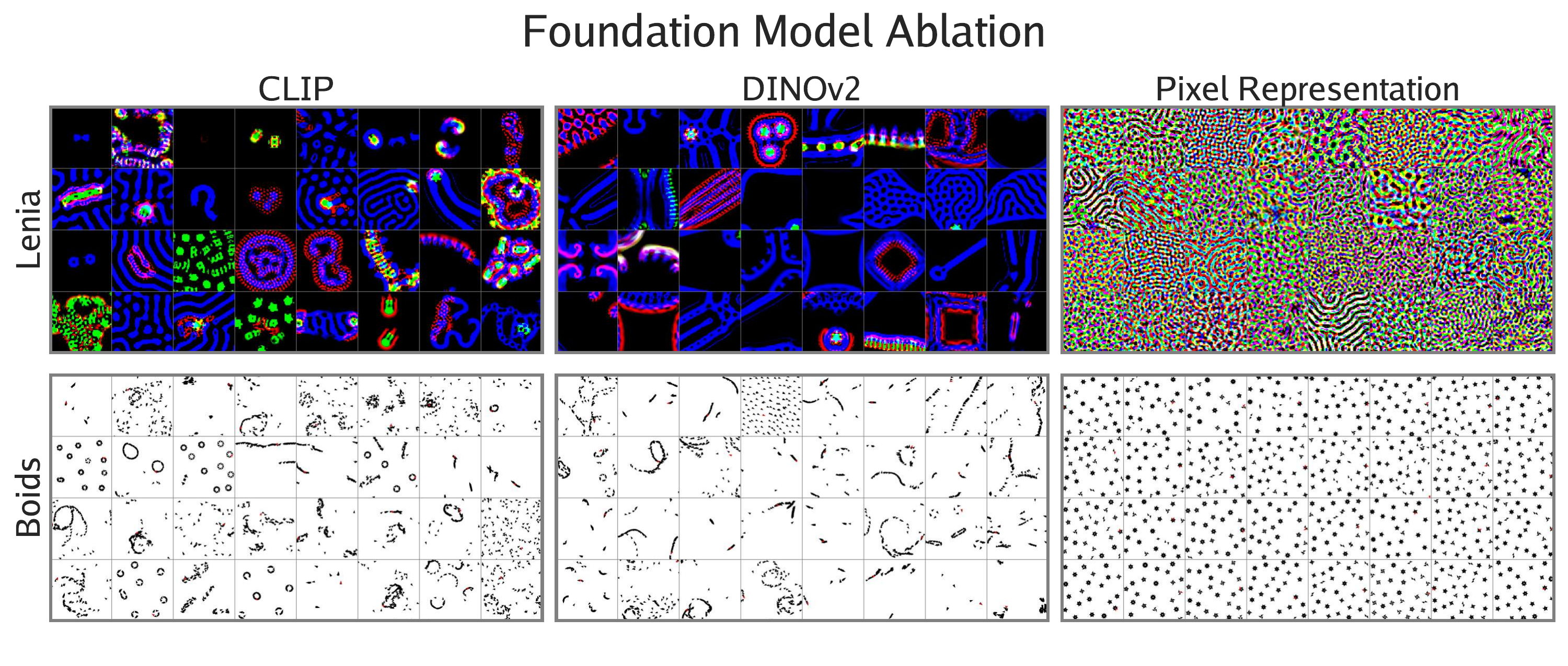}
  \vspace{-10mm}
  \caption{
        \textbf{Importance of Foundation Models:}
        The FM is ablated in the illumination experiments.
        CLIP seems to be slightly better than DINOv2 in creating human-aligned diversity, but both are significantly better than a pixel based representations.
  }
  \label{fig:fm_ablation}
\end{figure}

\subsection{Agnostic to Foundation Model}
To study the importance of using the proper representation space, we ablate the FM used during illumination of Lenia and Boids.
We keep the rest of the pipeline unchanged and replace CLIP with either DINOv2 or a low-level pixel-based representation, allowing us to isolate the precise impact of the FM.

As shown in Figure~\ref{fig:fm_ablation}, for producing human-aligned diversity, CLIP seems slightly better than DINOv2, but both are qualitatively better than the pixel representation.
This result highlights the importance of deep FM representations over low-level metrics when measuring human notions of diversity.

\vspace{-0.1in}
\section{Conclusion}
\vspace{-0.1in}
\paragraph{Summary}
This project launches a new paradigm in ALife by taking the first step towards using FMs to automate the search for interesting simulations.
Our approach is effective in finding target, open-ended, and diverse simulations over a wide spectrum of substrates.
Additionally, FMs enable the quantification of many qualitative phenomena in ALife, offering a path to replacing low-level complexity metrics with deep representations aligned with humans.

\vspace{-0.1in}
\paragraph{Discussion}
Because this project is agnostic to the FM and substrate used, it raises the question of which ones to use.
The choice of FMs seems to not matter much from our experiments, and FMs in general may also be converging to similar representations of reality~\citep{huh2024platonic}.

The proper substrate largely depends on the phenomena that is being studied (e.g.\ self-organization, open-ended evolution, etc.).
The most expressive substrate would simply parameterize all the RGB pixels of an entire video, but is useless for studying emergence.
The most insightful substrates \textit{bake in as little information as possible}, while \textit{maintaining vast emergent capabilities}.
For example, the periodic table of elements can be defined with little information, yet gives rise to the entirety of the observable universe.

Eventually, with the proper substrate, more powerful FMs, and enough compute, this paradigm may allow researchers to automatically search for worlds which start off as ``\texttt{simple cells in primordial soup}'', then undergo ``\texttt{a Cambrian explosion of complexity}'', and eventually become ``\texttt{an artificial alien civilization}.''
Researchers could alternatively search for hypothetical worlds where life evolves without DNA.
Finding open-ended worlds would solve one of ALife's grand challenges \citep{bedau2000open, stanley2017open}.
Illuminating such a substrate could help map the space of possible lifeforms and intelligences, giving a taxonomy of life as it could be in the computational universe.

This work can be generalized by replacing the image-language FM with video-language FMs that natively process the temporal nature of simulations~\citep{tang2023video, xu2021videoclip} or with 3-D FMs to handle 3-D simulations.
To leverage the recent advances of LLMs, images can be converted to text via image-to-text models, allowing all analyses to be done in text space.
Instead of searching for ALife simulations, a similar approach could be constructed for low-level physics research.
For example, in Wolfram's Physics Project \citep{wolfram2020class}, one could search for the hypergraph update rule which emerges structures that an FM considers natural.
At a meta-level, LLMs could be useful for generating code that describes the substrates themselves, driven by higher-level research agendas, similar to~\citet{faldor2024omni, lu2024discovering, lu2024ai}.

{
\paragraph{Acknowledgments }
We thank Ettore Randazzo for a discussion on the framing of the project.
This work was supported by an NSF GRFP Fellowship to A.K., a Packard Fellowship and a Sloan Research Fellowship to P.I., and by ONR MURI grant N00014-22-1-2740.
}

\bibliographystyle{plainnat}
\bibliography{references}

\newpage
\appendix

\section{Substrate Details}
\begin{itemize}
    \item
\textbf{Boids} simulates $N$ ``bird-like objects'' (boids) moving in a 2-D Euclidean space~\citep{reynolds1987flocks}.
All boids weight-share a single neural network that steers each boid left or right based on the $K$ nearby boids in its local frame of reference .
The substrate is the weight-space of the neural network.
For all experiments, we use $N=128$ boids and $K=16$ neighbors.
Each boid's neural network sees its boid neighbors' position and velocity in its own local frame of reference.
The position and velocity are processed separately for each neighbor, and eventually mean pooled to get a global vector, making the entire boid permutation invariant to its local neighbors index ordering.
This vector is then further processed to get a continuous control signal of how much to turn left or right.
The speed of all boids is fixed to ensure stability.
We simulate boids for a total of 1000 timesteps

    \item
\textbf{Particle Life} (or Clusters) simulates $N$ particles, each belonging to one of $K$ types, interacting in a 2-D Euclidean space \citep{ventrella2017clusters, mohr2023particlelife}.
The substrate is the space of $K\times K$ interaction matrices and the $\beta$ parameters to determine how close particles can get to each other.
The initial state is sampled randomly and the particles self-organize to form dynamic patterns.
For all experiments we use $N=5000$ particles and $K=6$ types of particles.
Following~\citep{mohr2023particlelife}, the interaction rule between particles is based on the $K\times K$ interaction matrix entry, where a positive value ensures attraction and a negative value ensures repulsion.
For each type of particle, there is an additional $\beta$ parameter which controls the distance at which the particle starts repelling all other particles, as to avoid excessive clumping. 
The asymmetry in the interaction matrix allows for complex interactions to take form.
Particle Life is simulated for a total of 1000 timesteps.

    \item
\textbf{Life-like Cellular Automata} (CA) generalizes Conway’s Game of Life to all binary-state CAs that operate in 2-D lattice where the state transition depends solely on the number of living Moore neighbors and the current state of the cell \citep{wojtowicz2024cellular}. 
The substrate has $2^{18}=262$,$144$ possible simulations.
The initial state is a random $64\times 64$ grid, where the sparsity level is uniformly sampled from $\mathcal{U}(0.05, 0.4)$.
Life-like CAs are simulated for 2048 timesteps.

    \item
\textbf{Lenia} generalizes Conway’s Game of Life to continuous space and time, allowing for higher dimensions, multiple kernels, and multiple channels~\citep{chan2018lenia}.
We utilize the LeniaBreeder codebase, which defines the substrate with $45$ dimensions for dynamics and $32 \times 32 \times 3 = 3$,$072$ dimensions for the initial state \citep{faldor2024toward}.
The search space is centered at a solution found by~\citet{chan2020lenia}.
To maintain stability, the dynamics parameters are bounded by an lower and upper threshold using a sigmoid function.
Lenia is simulated for 256 timesteps.
    \item
\textbf{Neural Cellular Automata} (NCA) parameterizes any continuous cellular automata by representing the local transition function through a neural network \citep{mordvintsev2020growing}.
The substrate is the weight-space of the neural network.
The architecture is similar to the original NCA architecture but replaces the fixed sobel filters with learned convolutional filters.
The initial state is randomly sampled by rendering a randomly sized circle at a random location.
We simulate the NCA for 256 timesteps.

\end{itemize}

\begin{figure}[h!]
    \centering
    \includegraphics[width=1.0\linewidth]{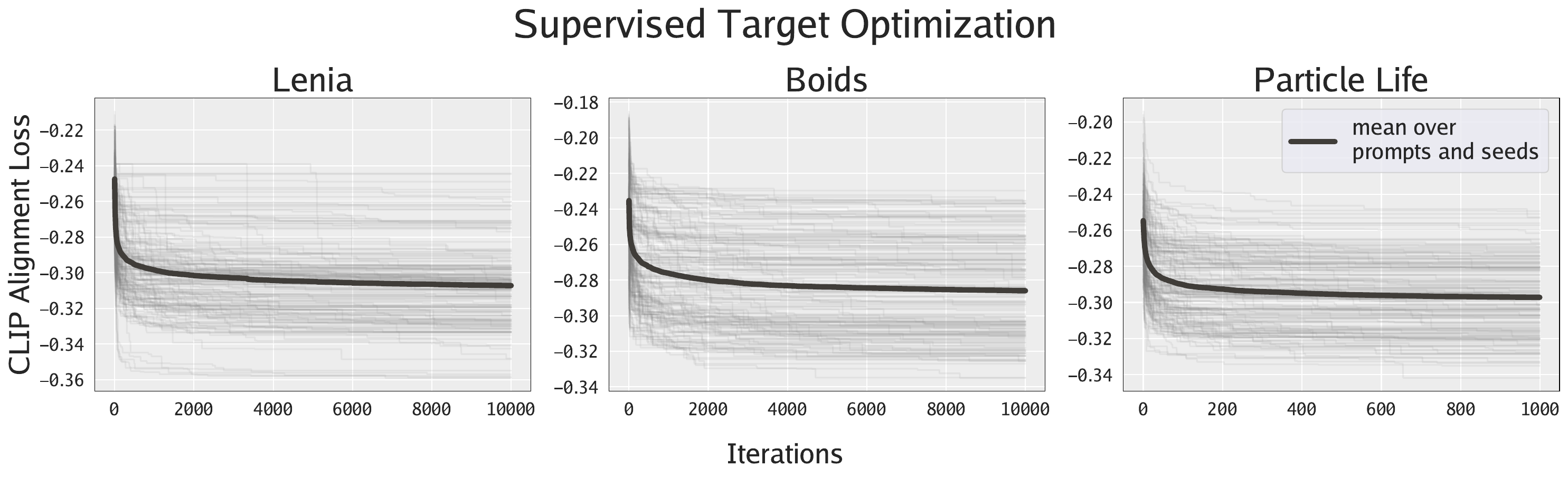}
    \caption{The training curve for the supervised target experiments.
    The y-axis measures the negative score from Equation~\ref{eq:supervised}.
    }
    \label{fig:supervised_learning_curve}
\end{figure}

\section{Experiment Details}
The code and instructions to replicate all experiments can be found at \href{https://github.com/SakanaAI/asal}{https://github.com/SakanaAI/asal}.
The codebase includes scripts that allow researchers to run ASAL in large-scale experiments, including hyperparameter sweeps.

\subsection{Target}

\paragraph{\textbf{Single Target}}
For single prompt optimization, the Sep-CMA-ES algorithm is employed.
Each simulation is evaluated using a single initial state seed.
The optimization runs with a population size of 16 for 10,000 iterations, using a mutation rate of 0.1.

\begin{figure}[p]
    \centering
    \includegraphics[width=1.0\linewidth]{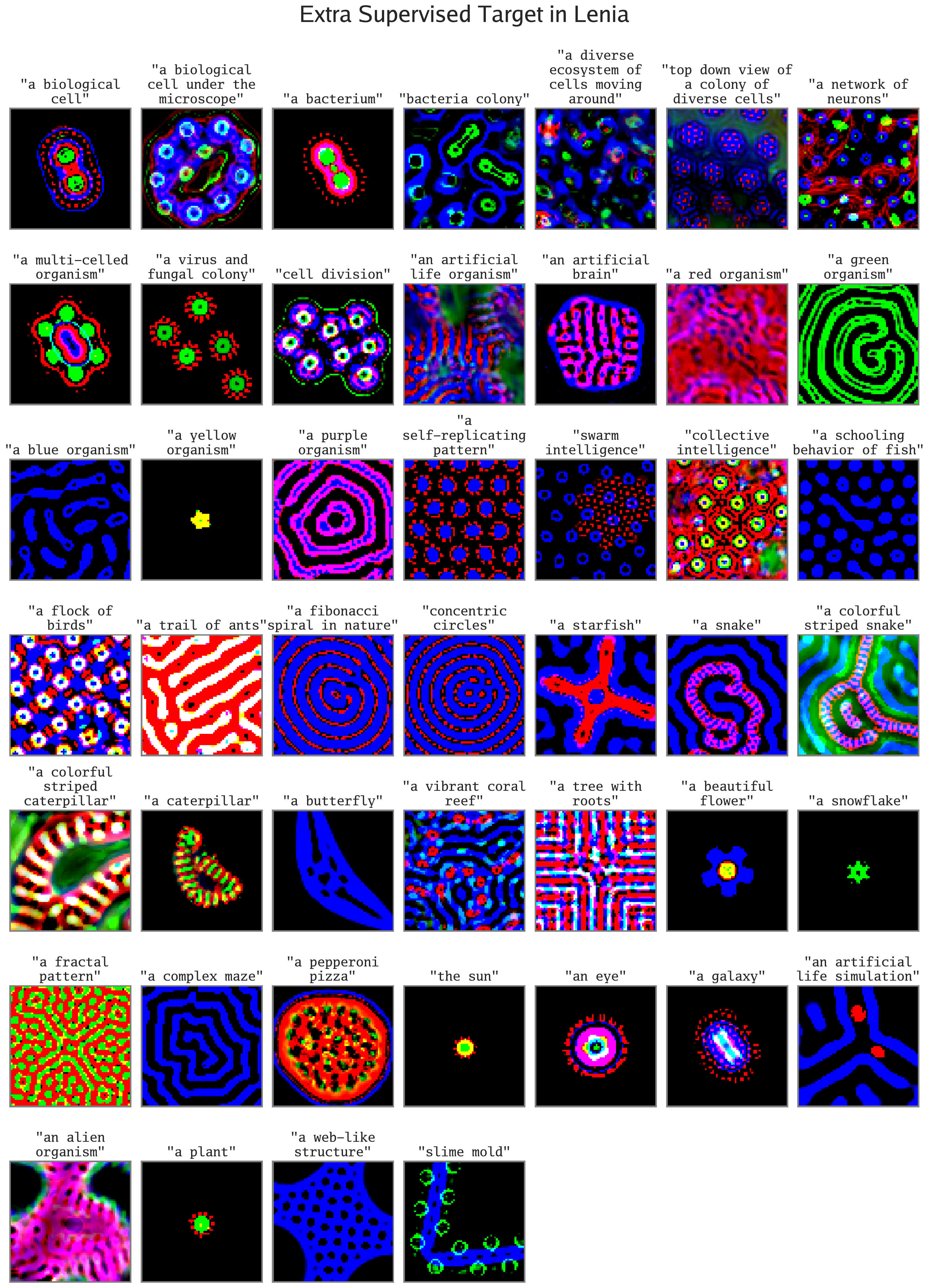}
    \caption{Extra supervised targets in Lenia.}
    \label{fig:supervised_extra_lenia}
\end{figure}

\begin{figure}[p]
    \centering
    \includegraphics[width=1.0\linewidth]{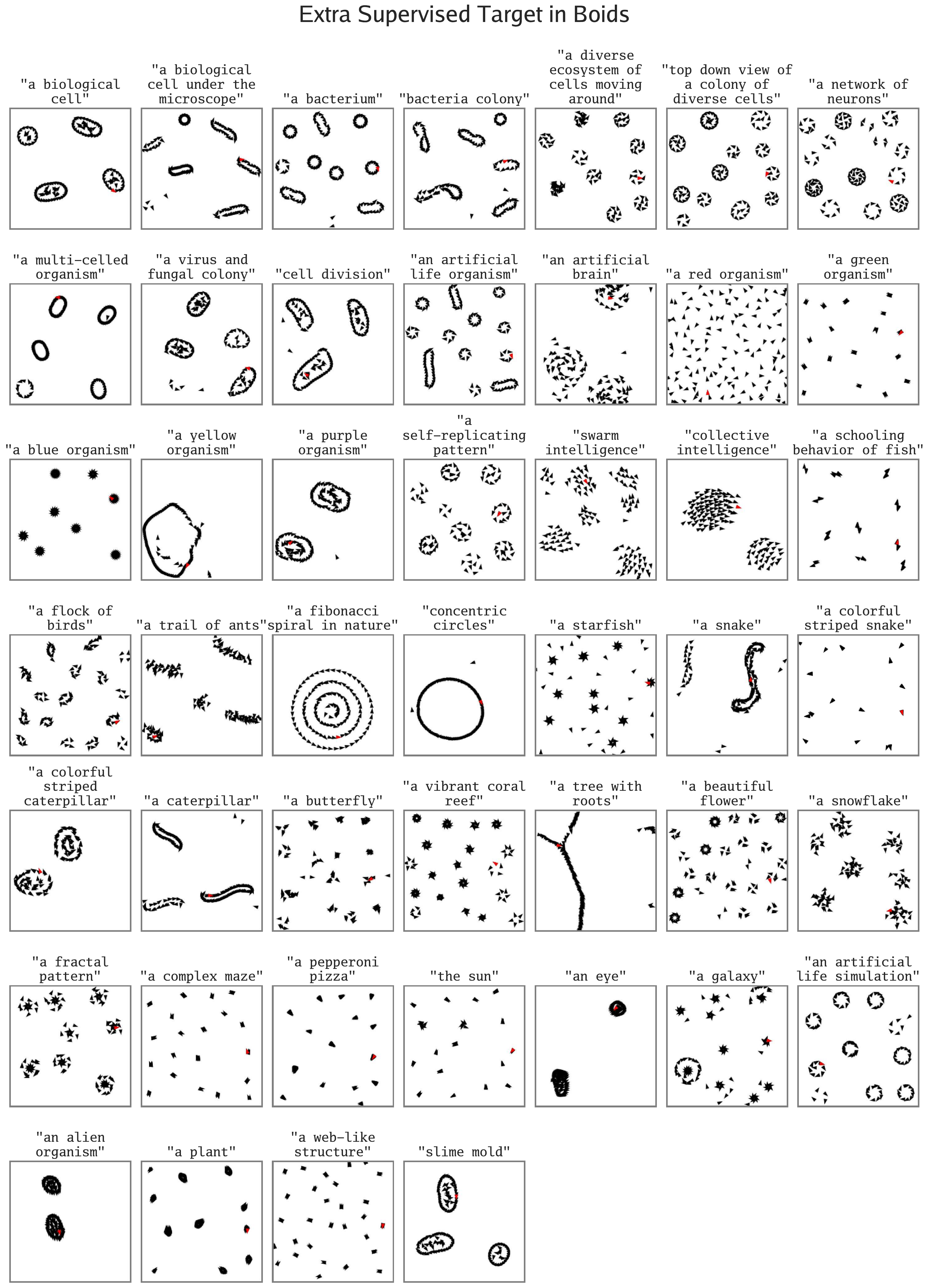}
    \caption{Extra supervised targets in Boids.}
    \label{fig:supervised_extra_boids}
\end{figure}

\begin{figure}[p]
    \centering
    \includegraphics[width=1.0\linewidth]{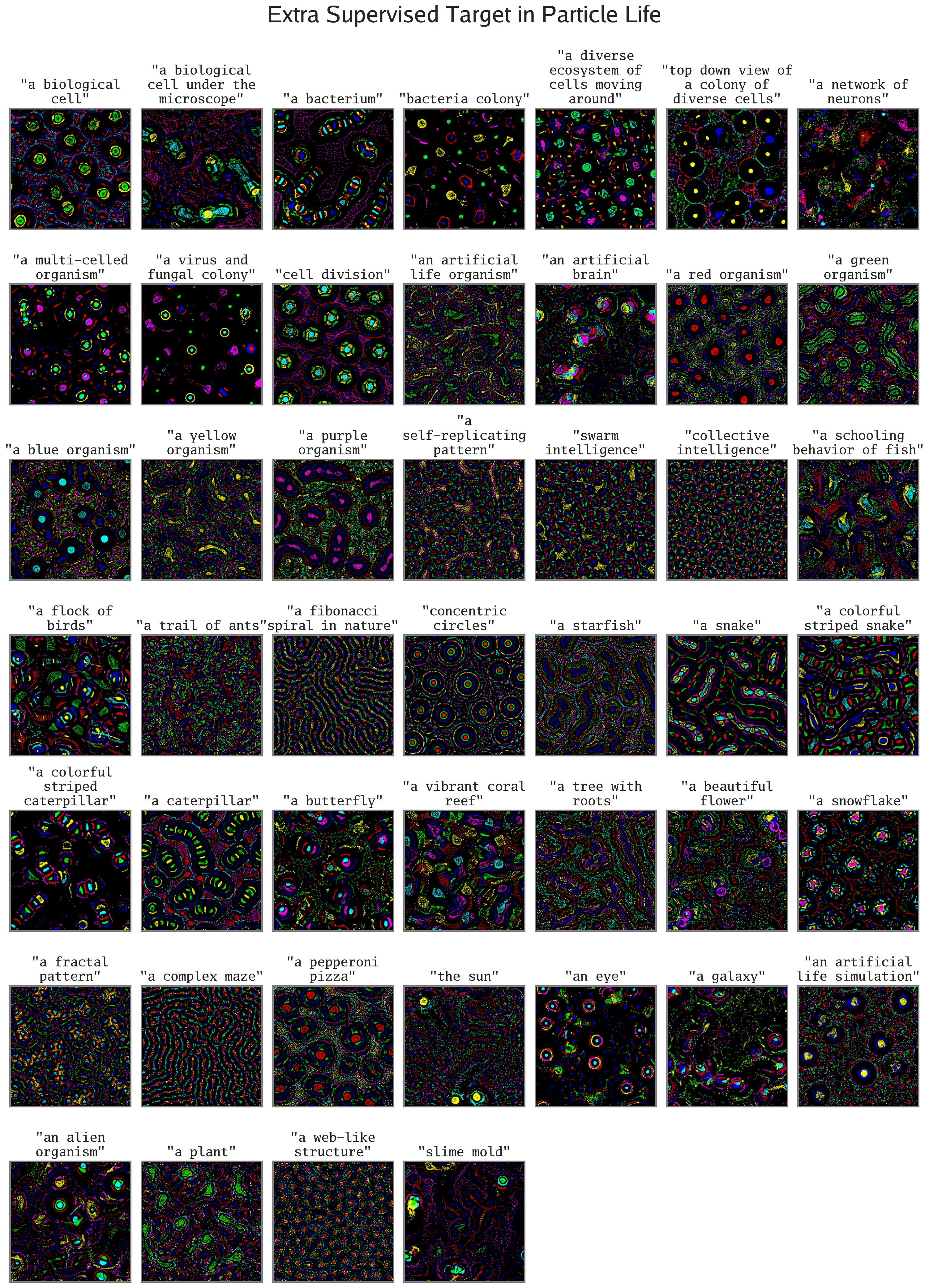}
    \caption{Extra supervised targets in Particle Life.}
    \label{fig:supervised_extra_plife}
\end{figure}

The results from additional prompts are shown in Figure~\ref{fig:supervised_extra_lenia},
Figure~\ref{fig:supervised_extra_boids}, and
Figure~\ref{fig:supervised_extra_plife}.
The training curves for all experiments are shown in Figure~\ref{fig:supervised_learning_curve}.

\paragraph{\textbf{Temporal Targets}}
For the temporal target experiments in the NCA substrate, the Adam optimizer with truncated backpropagation through time is used.
Following the original NCA paper~\citep{mordvintsev2020growing}, a pool of 256 simulation states is maintained.
Batches of 8 states are sampled and rolled out 16 timesteps at a time, where truncated backpropagation through time is applied.
The learning rate is set to 3e-4, and optimization runs for 1,000,000 iterations with gradient clipping set to 1.

\begin{figure}[th!]
    \centering
    \includegraphics[width=1.0\linewidth]{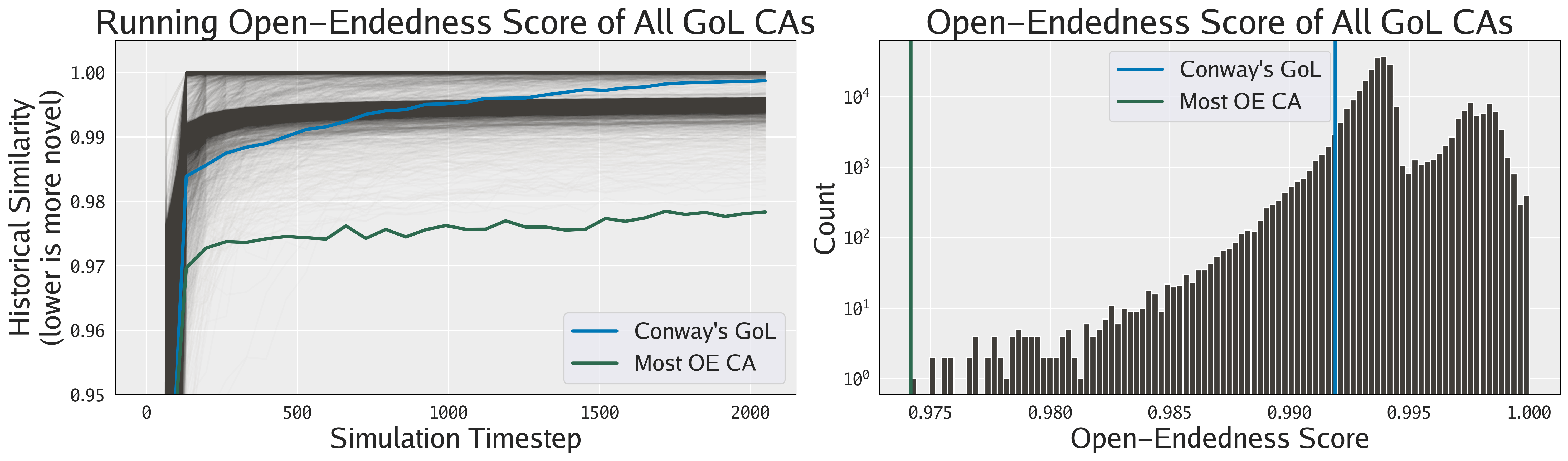}
    \caption{The open-endedness score from Equation~\ref{eq:oe} for all the life-like CAs.
    The left subplot shows how the open-endedness score changes over the simulation time.
    The right subplot is a histogram of the average open-endedness score of all the CAs.}
    \label{fig:oe_gol_learning_curve}
\end{figure}
\begin{figure}[th!]
    \centering
    \includegraphics[width=1.0\linewidth]{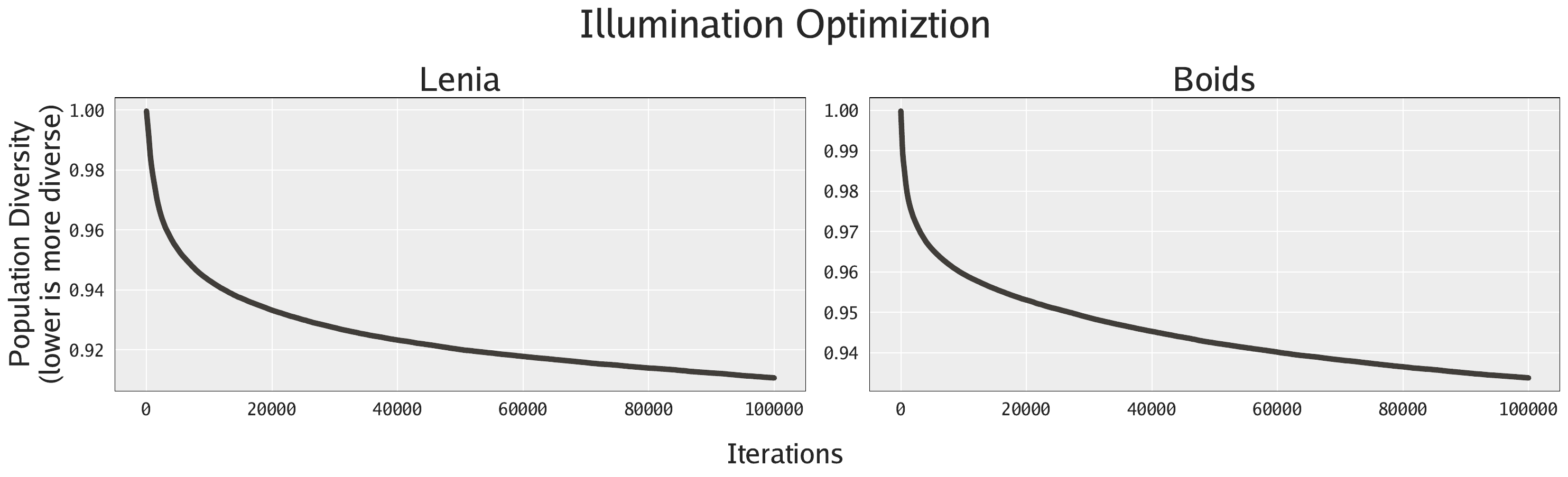}
    \caption{The training curve for the illumination experiments.
    The y-axis measures the diversity of the population of solutions as measured by Equation~\ref{eq:illumination}.
    }
    \label{fig:illumination_learning_curve}
\end{figure}

\subsection{Open-Endedness}

For the open-endedness experiments, brute force search is applied over the 262,144 simulations.
Each simulation runs with 256 random initial states for 2,048 timesteps.
The open-endedness metric is calculated by subsampling 32 timesteps for use in Equation~\ref{eq:oe}.

\begin{figure}[p]
  \centering
  \includegraphics[width=1.0\linewidth]{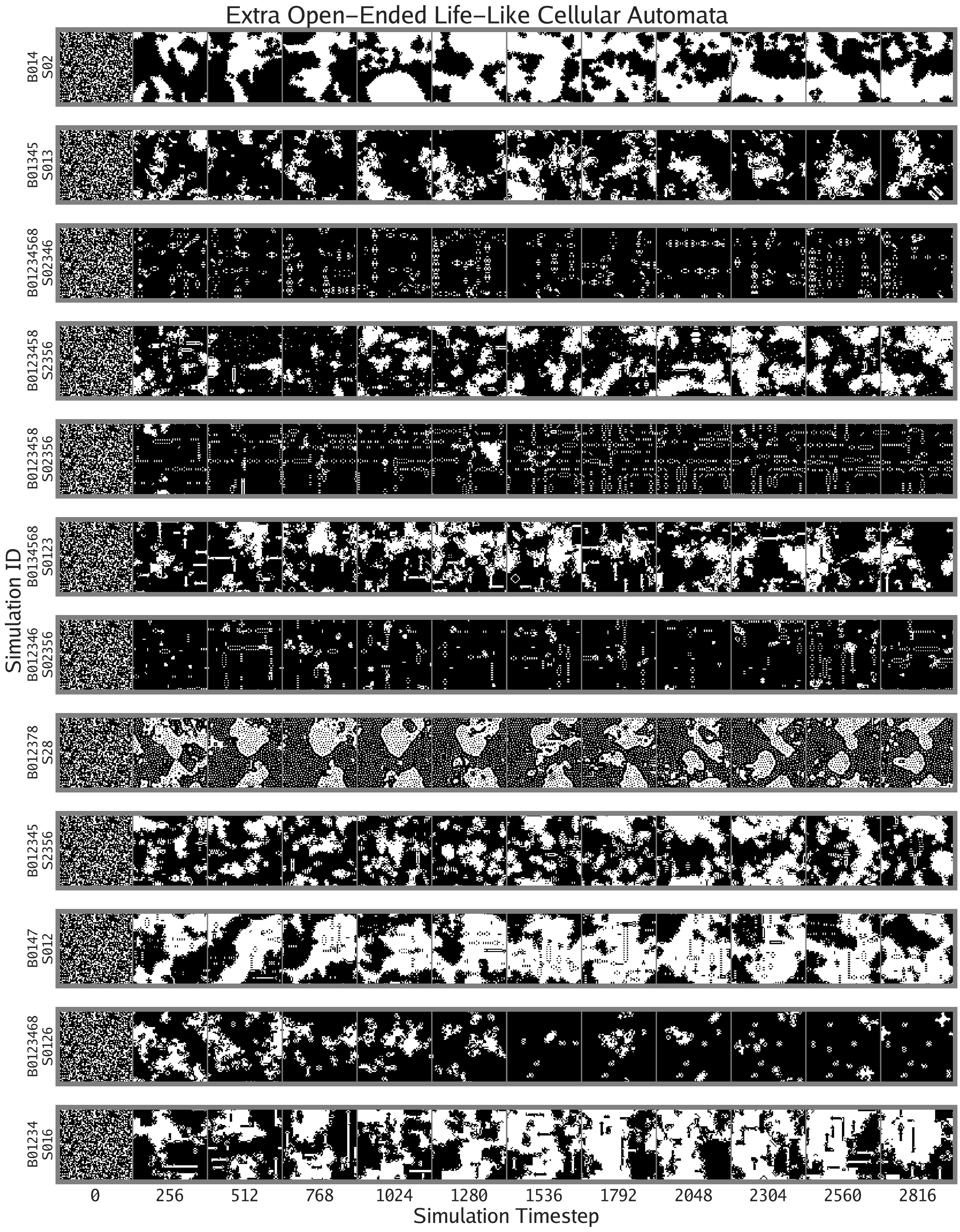}
  \caption{
  }
  \label{fig:oe_gol_extra}
\end{figure}

Aside from the CAs shown in the main paper, additional open-ended CAs are visualized in Figure~\ref{fig:oe_gol_extra}.
The information gathered from the brute force search is shown in Figure~\ref{fig:oe_gol_learning_curve}.

\subsection{Illumination}

For illumination, a custom genetic algorithm is used for the search.
The algorithm maintains a population of 8,192 solutions initialized as all zeros.
In each iteration, 32 solutions are sampled and mutated by adding a random Gaussian vector with a standard deviation of 0.1.
Afterward, the population grows to 8,224 solutions, and the least novel solution is removed 32 times, reducing the population back to 8,192.
The least novel solution is identified as the one with the smallest average distance to its two nearest neighbors.
This process is repeated for 100,000 iterations.

\begin{figure}[p]
    \centering
    \rotatebox{-90}{\includegraphics[width=\textheight,keepaspectratio]{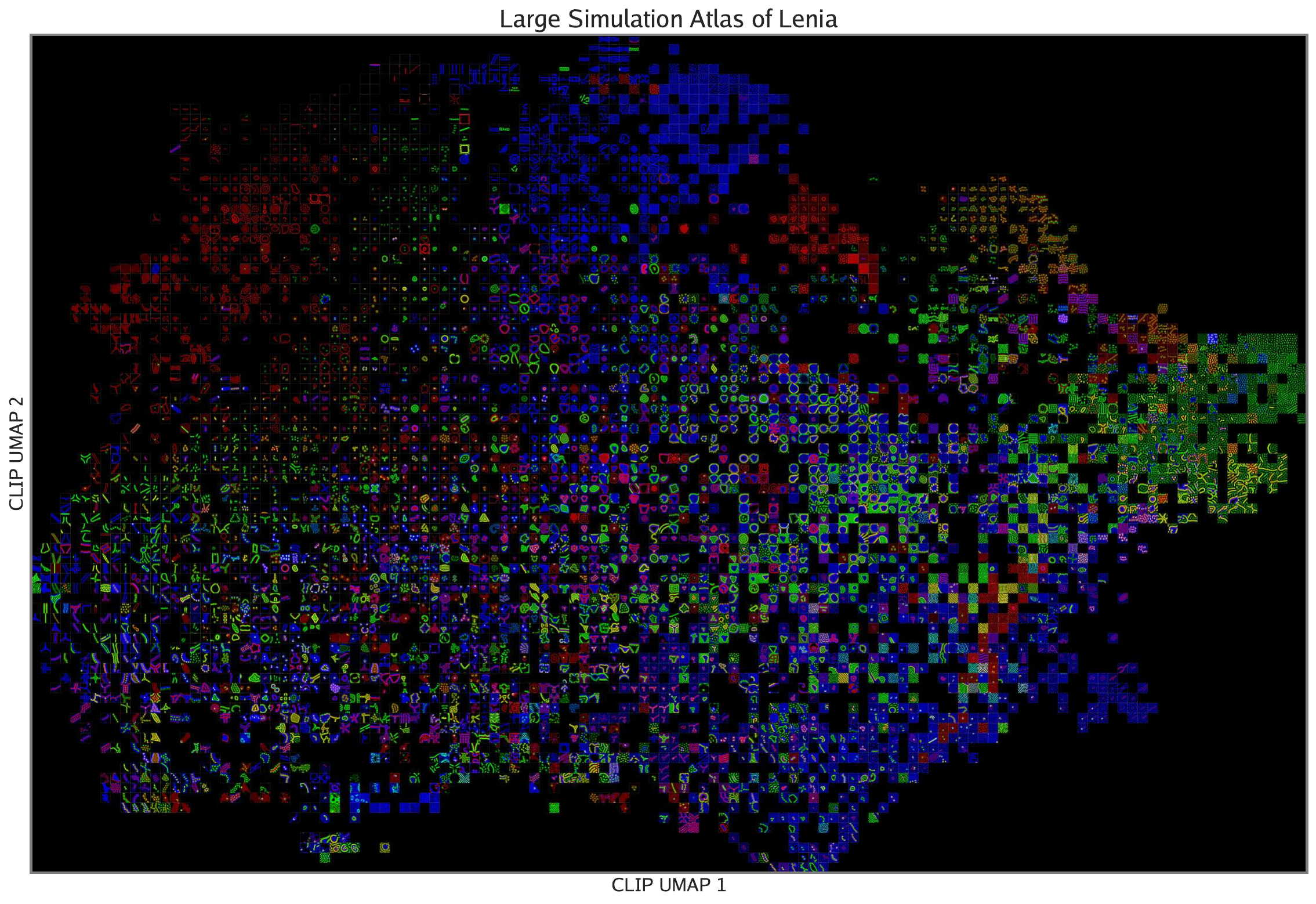}}
    \caption{Large Simulation Atlas for Lenia.}
    \label{fig:atlas_lenia_large}
\end{figure}

\begin{figure}[p]
    \centering
    \rotatebox{-90}{\includegraphics[width=\textheight,keepaspectratio]{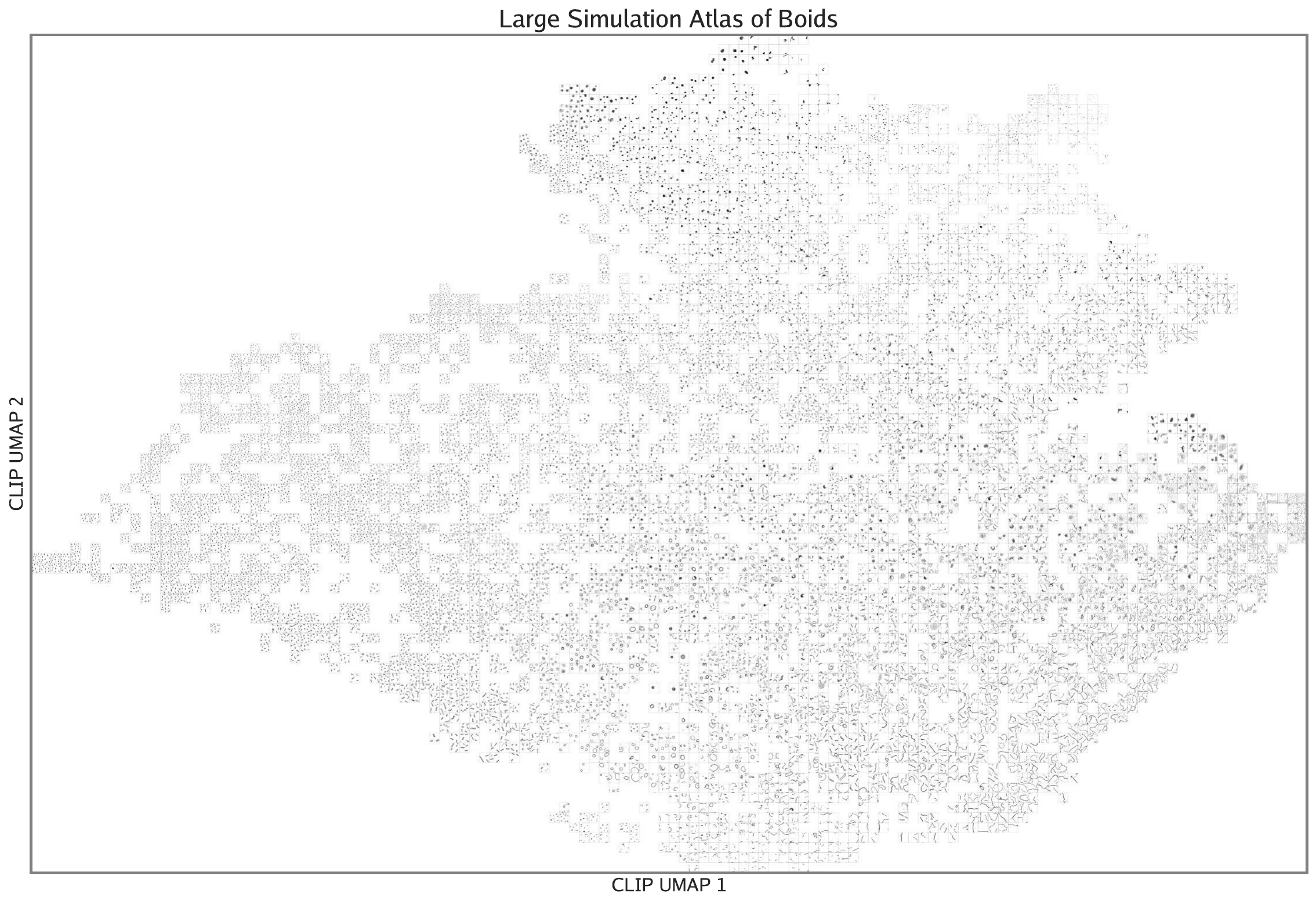}}
    \caption{Large Simulation Atlas for Boids.}
    \label{fig:atlas_boids_large}
\end{figure}

\begin{figure}[p]
    \centering
    \rotatebox{-90}{\includegraphics[width=\textheight,keepaspectratio]{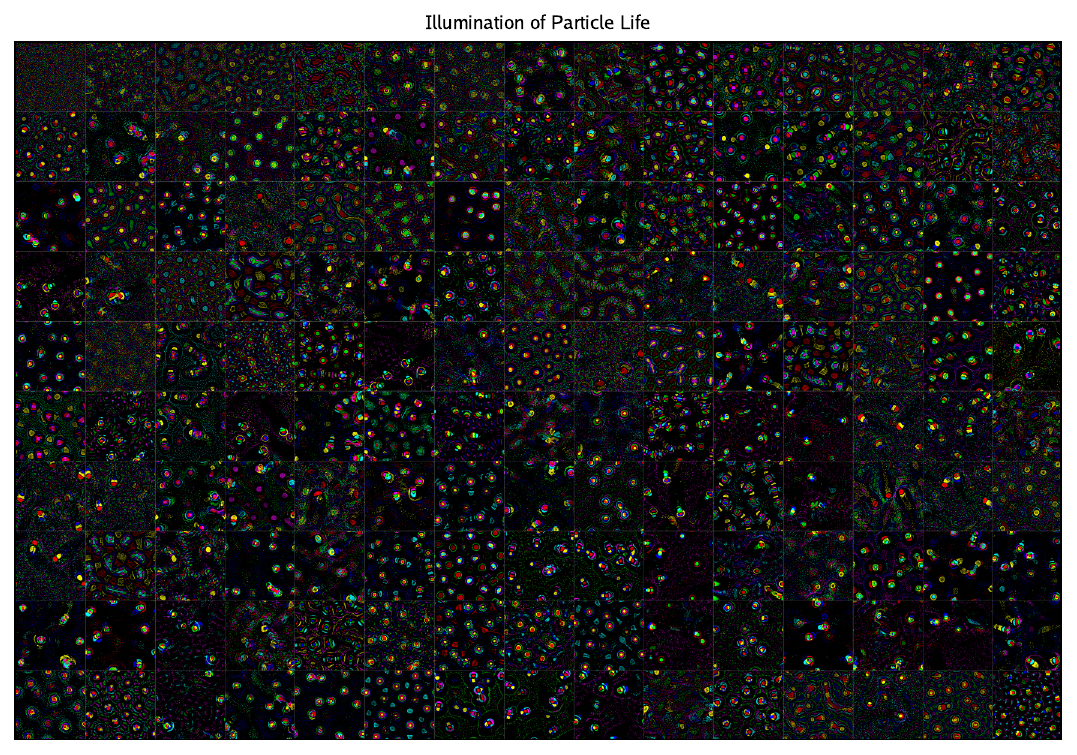}}
    \caption{Illumination of Particle Life.}
    \label{fig:illumination_plife}
\end{figure}

Larger simulation atlases are shown in Figure~\ref{fig:atlas_lenia_large} and Figure~\ref{fig:atlas_boids_large}.
The training curves for these experiments are shown in Figure~\ref{fig:illumination_learning_curve}.
The illumination of Particle Life is shown in Figure~\ref{fig:illumination_plife}.

\newpage

\begin{figure}[p]
    \centering
    \includegraphics[width=1.0\textwidth,keepaspectratio]{figs/atlas_gol_0.png}
    \includegraphics[width=1.0\textwidth,keepaspectratio]{figs/atlas_gol_1.png}
    \includegraphics[width=1.0\textwidth,keepaspectratio]{figs/atlas_gol_2.png}
    \caption{
    Simulation Atlases for the Life-Like CA substrate.
    Top: Atlas of the main cluster from Figure~\ref{fig:oe_gol_3}.
    Middle: Atlas of the bottom-left cluster from Figure~\ref{fig:oe_gol_3}.
    Bottom: Atlas of the bottom-right cluster from Figure~\ref{fig:oe_gol_3}.
    }
    \label{fig:atlas_gol}
\end{figure}

\end{document}